%% file: arxiv.tex
\documentclass[letterpaper, 10 pt, conference]{ieeeconf}  

\IEEEoverridecommandlockouts                              
\usepackage{graphicx}
\graphicspath{{icra/}{icra/figures/}}
\usepackage[dvipsnames,svgnames,table]{xcolor}
\usepackage[utf8]{inputenc}
\usepackage{amsmath,amssymb}
\usepackage{dsfont}
\usepackage{tabularx}
\usepackage{booktabs}
\usepackage{multirow}
\usepackage{makecell}
\usepackage{adjustbox}
\usepackage{float}
\usepackage{soul}
\let\labelindent\relax   
\usepackage{enumitem}
\usepackage{url}
\usepackage{capt-of}
\usepackage{tcolorbox}   
\tcbuselibrary{listings,breakable}
\usepackage[hidelinks]{hyperref}
\definecolor{linkblue}{RGB}{0,84,166}

\makeatletter
\def\bstctlcite{\@ifnextchar[{\@bstctlcite}{\@bstctlcite[@auxout]}}
\def\@bstctlcite[#1]#2{\@bsphack
  \@for\@citeb:=#2\do{%
    \edef\@citeb{\expandafter\@firstofone\@citeb}%
    \if@filesw\immediate\write\csname #1\endcsname{\string\citation{\@citeb}}\fi}%
  \@esphack}
\makeatother

\newcommand{\ourmethod}{ProTracer}
\newcommand{\ourdataset}{FailTime}

\usepackage{marvosym} 

\title{\LARGE \bf
\ourmethod{}: Proprioception-Guided Failure Diagnosis in \\ Robot Manipulation
}

\author{Anonymous Author(s)%
}

\author{Chang Dong$^{1}$, Mehdi Hosseinzadeh$^{1}$, King Hang Wong$^{1}$, Lingqiao Liu$^{1}$,\\ Francois Fraysse$^{2}$, Feras Dayoub$^{1}$, and Minh Hoai$^{1,\dagger}$%
\\[3pt]{\normalsize\href{https://protracer-failure.github.io/}{\textcolor{linkblue}{https://protracer-failure.github.io}}}%
\thanks{$^{\dagger}$Corresponding author.}%
\thanks{$^{1}$Australian Institute for Machine Learning and $^{2}$School of Allied Health and Human Performance, Adelaide University, Australia.}%
\thanks{Emails: \href{mailto:chang.dong@adelaide.edu.au}{Chang Dong}, \href{mailto:mehdi.hosseinzadeh@adelaide.edu.au}{Mehdi Hosseinzadeh}, \href{mailto:kinghang.wong@adelaide.edu.au}{King Hang Wong}, \href{mailto:lingqiao.liu@adelaide.edu.au}{Lingqiao Liu}, \href{mailto:francois.fraysse@adelaide.edu.au}{Francois Fraysse}, \href{mailto:feras.dayoub@adelaide.edu.au}{Feras Dayoub}, \href{mailto:minhhoai.nguyen@adelaide.edu.au}{Minh Hoai}.}%
}

\begin{document}
\input{icra/definitions}
\bstctlcite{IEEEexample:BSTcontrol}   

\setlength{\textfloatsep}{6pt plus 2pt minus 2pt}
\setlength{\dbltextfloatsep}{6pt plus 2pt minus 2pt}
\setlength{\floatsep}{6pt plus 2pt minus 2pt}
\setlength{\dblfloatsep}{6pt plus 2pt minus 2pt}
\setlength{\abovecaptionskip}{3pt}
\setlength{\belowcaptionskip}{0pt}
\maketitle
\thispagestyle{empty}
\pagestyle{empty}

\input{icra/sections/abstract}

\input{icra/sections/intro}

\input{icra/sections/related}
\input{icra/sections/method}
\input{icra/sections/experiments}
\input{icra/sections/conclusion}


\section*{ACKNOWLEDGMENT}
The authors sincerely thank the additional annotators, Yichao Cai, Shuying Piao, and Lishan Yang, for their valuable contributions to the annotation and quality control of the \ourdataset{} benchmark. Their careful review and feedback helped improve the consistency and reliability of the annotations.

\bibliographystyle{IEEEtran}
\bibliography{icra/references}   

\newif\ifappendix
\appendixfalse
\ifappendix
\clearpage
\section*{APPENDIX}
\renewcommand{\thesection}{\Alph{section}}\setcounter{section}{0}   
\renewcommand{\thesubsection}{\thesection.\arabic{subsection}}
\input{icra/appendix/analysis}
\input{icra/appendix/taxonomy}
\input{icra/appendix/dataset}

\input{icra/appendix/method}
\input{icra/appendix/evaluation}
\fi

\end{document}

%% file: icra/definitions.tex
\definecolor{ourshl}{RGB}{230,240,250}
\def\mA{\mathcal{A}}
\def\mB{\mathcal{B}}
\def\mC{\mathcal{C}}
\def\mD{\mathcal{D}}
\def\mE{\mathcal{E}}
\def\mF{\mathcal{F}}
\def\mG{\mathcal{G}}
\def\mH{\mathcal{H}}
\def\mI{\mathcal{I}}
\def\mJ{\mathcal{J}}
\def\mK{\mathcal{K}}
\def\mL{\mathcal{L}}
\def\mM{\mathcal{M}}
\def\mN{\mathcal{N}}
\def\mO{\mathcal{O}}
\def\mP{\mathcal{P}}
\def\mQ{\mathcal{Q}}
\def\mR{\mathcal{R}}
\def\mS{\mathcal{S}}
\def\mT{\mathcal{T}}
\def\mU{\mathcal{U}}
\def\mV{\mathcal{V}}
\def\mW{\mathcal{W}}
\def\mX{\mathcal{X}}
\def\mY{\mathcal{Y}}
\def\mZ{\mathcal{Z}} 

\def\bbN{\mathbb{N}} 
\def\bbR{\mathbb{R}} 
\def\bbP{\mathbb{P}} 
\def\bbQ{\mathbb{Q}} 
\def\bbE{\mathbb{E}}

\def\1n{\mathbf{1}_n}
\def\0{\mathbf{0}}
\def\1{\mathbf{1}}

\def\A{{\bf A}}
\def\B{{\bf B}}
\def\C{{\bf C}}
\def\D{{\bf D}}
\def\E{{\bf E}}
\def\F{{\bf F}}
\def\G{{\bf G}}
\def\H{{\bf H}}
\def\I{{\bf I}}
\def\J{{\bf J}}
\def\K{{\bf K}}
\def\L{{\bf L}}
\def\M{{\bf M}}
\def\N{{\bf N}}
\def\O{{\bf O}}
\def\P{{\bf P}}
\def\Q{{\bf Q}}
\def\R{{\bf R}}
\def\S{{\bf S}}
\def\T{{\bf T}}
\def\U{{\bf U}}
\def\V{{\bf V}}
\def\W{{\bf W}}
\def\X{{\bf X}}
\def\Y{{\bf Y}}
\def\Z{{\bf Z}}

\def\a{{\bf a}}
\def\b{{\bf b}}
\def\c{{\bf c}}
\def\d{{\bf d}}
\def\e{{\bf e}}
\def\f{{\bf f}}
\def\g{{\bf g}}
\def\h{{\bf h}}
\def\i{{\bf i}}
\def\j{{\bf j}}
\def\k{{\bf k}}
\def\l{{\bf l}}
\def\m{{\bf m}}
\def\n{{\bf n}}
\def\o{{\bf o}}
\def\p{{\bf p}}
\def\q{{\bf q}}
\def\r{{\bf r}}
\def\s{{\bf s}}
\def\t{{\bf t}}
\def\u{{\bf u}}
\def\v{{\bf v}}
\def\w{{\bf w}}
\def\x{{\bf x}}
\def\y{{\bf y}}
\def\z{{\bf z}}

\def\balpha{\mbox{\boldmath{$\alpha$}}}
\def\bbeta{\mbox{\boldmath{$\beta$}}}
\def\bdelta{\mbox{\boldmath{$\delta$}}}
\def\bgamma{\mbox{\boldmath{$\gamma$}}}
\def\blambda{\mbox{\boldmath{$\lambda$}}}
\def\bsigma{\mbox{\boldmath{$\sigma$}}}
\def\btheta{\mbox{\boldmath{$\theta$}}}
\def\bomega{\mbox{\boldmath{$\omega$}}}
\def\bxi{\mbox{\boldmath{$\xi$}}}
\def\bnu{\mbox{\boldmath{$\nu$}}}                                  
\def\bphi{\mbox{\boldmath{$\phi$}}}
\def\bmu{\mbox{\boldmath{$\mu$}}}

\def\bDelta{\mbox{\boldmath{$\Delta$}}}
\def\bOmega{\mbox{\boldmath{$\Omega$}}}
\def\bPhi{\mbox{\boldmath{$\Phi$}}}
\def\bLambda{\mbox{\boldmath{$\Lambda$}}}
\def\bSigma{\mbox{\boldmath{$\Sigma$}}}
\def\bGamma{\mbox{\boldmath{$\Gamma$}}}
                                  
\newcommand{\myprob}[1]{\mathop{\mathbb{P}}_{#1}}

\newcommand{\myexp}[1]{\mathop{\mathbb{E}}_{#1}}

\newcommand{\mydelta}[1]{1_{#1}}

\newcommand{\myminimum}[1]{\mathop{\textrm{minimum}}_{#1}}
\newcommand{\mymaximum}[1]{\mathop{\textrm{maximum}}_{#1}}    
\newcommand{\mymin}[1]{\mathop{\textrm{minimize}}_{#1}}
\newcommand{\mymax}[1]{\mathop{\textrm{maximize}}_{#1}}
\newcommand{\mymins}[1]{\mathop{\textrm{min.}}_{#1}}
\newcommand{\mymaxs}[1]{\mathop{\textrm{max.}}_{#1}}  
\newcommand{\myargmin}[1]{\mathop{\textrm{argmin}}_{#1}} 
\newcommand{\myargmax}[1]{\mathop{\textrm{argmax}}_{#1}} 
\newcommand{\myst}{\textrm{s.t. }}

\newcommand{\denselist}{\itemsep -1pt}
\newcommand{\sparselist}{\itemsep 1pt}





\def\changemargin#1#2{\list{}{\rightmargin#2\leftmargin#1}\item[]}
\let\endchangemargin=\endlist
                                               
\newcommand{\cm}[1]{}

\newcommand{\mhoai}[1]{{\color{magenta}\textbf{[MH: #1]}}}
\newcommand{\FD}[1]{{\color{magenta}\textbf{[feras: #1]}}}

\newcommand{\mtodo}[1]{{\color{red}$\blacksquare$\textbf{[TODO: #1]}}}
\newcommand{\myheading}[1]{\vspace{0.5ex}\noindent \textbf{#1}}
\newcommand{\htimesw}[2]{\mbox{$#1$$\times$$#2$}}


%
%
%

\newcommand{\Sref}[1]{Sec.~\ref{#1}}
\newcommand{\Eref}[1]{Eq.~(\ref{#1})}
\newcommand{\Fref}[1]{Fig.~\ref{#1}}
\newcommand{\Tref}[1]{Table~\ref{#1}}

%% file: icra/sections/abstract.tex
\begin{abstract}
This paper presents a comprehensive framework for robot manipulation failure analysis that includes binary failure detection, failure categorization, explanation generation, and the additional capability of \emph{failure onset localization}, which aims to identify the earliest moment at which a robot execution deviates from a valid task-completion trajectory and is ultimately followed by task failure. To address these tasks, we propose \ourmethod{}, a training-free framework that leverages existing Vision-Language Models (VLMs) together with proprioceptive signals for failure analysis. Our method uses proprioceptive dynamics to identify temporally informative action boundaries and converts richer robot-state signals into structured natural-language descriptions that can be jointly analyzed together with visual observations by the VLM. This design combines the temporal precision of proprioceptive signals with the multimodal reasoning capabilities of modern VLMs without requiring additional model training. 
We further introduce \emph{FailTime}, a benchmark with synchronized visual and proprioceptive observations for evaluating conventional failure diagnosis tasks as well as failure onset localization. 
Experiments demonstrate that \ourmethod{} achieves strong performance across both conventional failure diagnosis tasks and the newly introduced failure onset localization task, highlighting the importance of proprioceptive reasoning for fine-grained temporal failure analysis.

\end{abstract}


%% file: icra/sections/intro.tex
\section{Introduction}

\label{sec:intro}


Failure diagnosis is important for robot learning and deployment, supporting applications such as data filtering, policy improvement, recovery planning, and safety analysis. However, existing approaches focus primarily on \emph{whether} and \emph{what} went wrong—detecting rollout failures, categorizing failure types, or explaining failure outcomes~\cite{liu2023reflect, duan2024aha, robofac2025, kite, vifailback}—while largely overlooking \emph{when} the failure first occurred. This distinction is important because manipulation failures may occur well before the end of a complete rollout, with the robot continuing to execute for many seconds after an initial mishap. Localizing failure onset enables more precise temporal credit assignment, recovery of successful trajectory prefixes from failed demonstrations, and finer-grained analysis of robot behavior.

\begin{figure}[!t]
  \centering
  \includegraphics[width=\columnwidth]{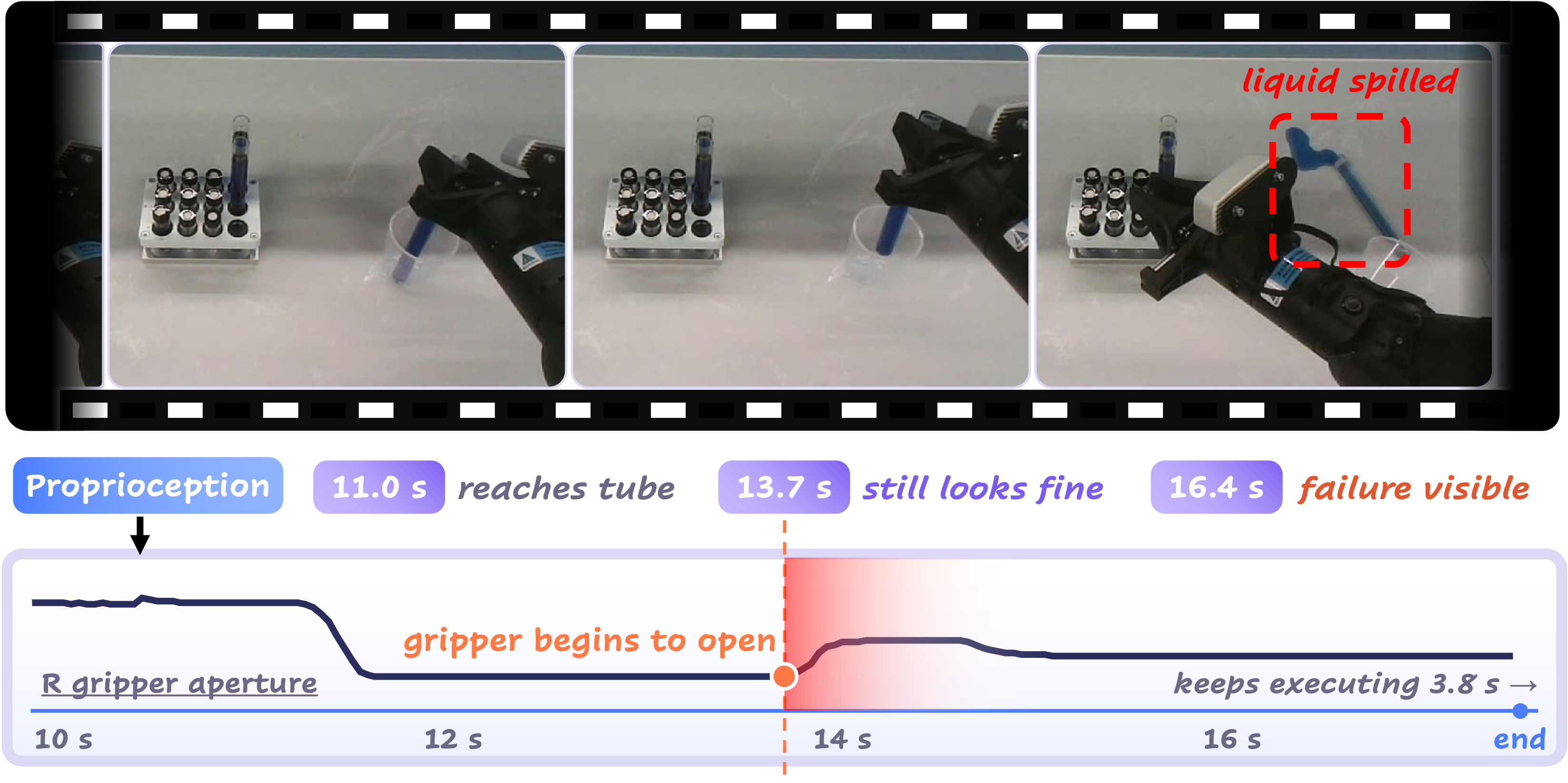}
  \caption{\textbf{Failure onset localization.} The gripper releases prematurely at 13.7\,s while the scene still looks normal; the spill only becomes visible at 16.4\,s, and the robot keeps executing afterwards. \ourmethod{} localizes this onset by using proprioception to pinpoint when the failure begins, and vision to identify what went wrong.}
  \label{fig:onset_teaser}
\end{figure}

\input{icra/figures/teaser_title}



We therefore study failure diagnosis as a joint problem that retains the existing diagnostic tasks and extends them to capture \emph{when} a failure begins. Specifically, we consider failure detection, failure categorization, and explanation generation, together with \emph{failure onset localization}. We define failure onset localization as the task of retrospectively identifying, given a completed manipulation rollout, the earliest moment at which the execution begins to lead to eventual task failure. 
As illustrated in Fig.~\ref{fig:onset_teaser}, the failure onset may be the moment the gripper releases prematurely, rather than a later moment when the consequence becomes visually apparent or the task is ultimately declared unsuccessful.



Importantly, this task cannot be adequately addressed by existing methods for online error or anomaly detection~\cite{agia2024unpacking,gu2025safe,romer2025failure,xu2025detect,fidel2026}. First, online methods operate only on observations available up to the current time, where early failure signals can be inherently ambiguous: apparent errors may later be recovered, while subtle errors may become recognizable only through their subsequent consequences. Consequently, the first alarm of an online detector is an unreliable onset estimate: a low threshold fires on transient deviations, a high one only after the failure is already visible, and neither marks where the execution actually went wrong. Second, online methods often rely on task-specific models of nominal execution, such as expected trajectories, task graphs, or learned distributions of successful behavior, requiring substantial task-specific data and training. In contrast, retrospective analysis can leverage the complete rollout and its known outcome to disambiguate earlier events without explicitly modeling every valid execution. Furthermore, access to the complete rollout enables joint reasoning over observations before and after a candidate failure onset, making it possible to leverage the multimodal reasoning capabilities of pretrained VLMs for training-free failure analysis.

Despite this potential, leveraging VLMs for failure onset localization is non-trivial, as the critical transitions associated with failure onset are often visually subtle and temporally brief. Dense video prompting can overwhelm VLMs with redundant observations~\cite{duan2024aha,robofac2025,vifailback}, while sparse visual sampling risks missing short, contact-rich transitions~\cite{kite}. Our key observation is that manipulation failures are not purely visual phenomena, but are also reflected in the robot’s execution dynamics. Gripper states and end-effector motion provide temporally precise cues to action transitions that can be difficult to infer from video alone, and such robot-state signals are routinely available in manipulation benchmarks and large-scale robot datasets~\cite{james2020rlbench,mu2021maniskill,pumacay2024colosseum,khazatsky2024droid,padalkar2023open}. Proprioceptive signals can therefore complement the visual reasoning capabilities of VLMs, providing fine-grained temporal cues for failure onset localization while also benefiting conventional failure diagnosis.


Technically, realizing this complementarity requires integrating proprioceptive signals into VLM-based reasoning. VLMs offer strong visual-language reasoning and semantic knowledge, but are not designed to interpret dense arrays of raw proprioceptive values. Training a new multimodal foundation model over vision, language, and robot-state trajectories is also impractical given the computational cost and limited annotated failure data. To address these challenges, we propose \ourmethod{}, a training-free framework that integrates proprioception into existing VLMs in two complementary ways. First, proprioceptive dynamics identify temporal boundaries corresponding to meaningful action transitions, focusing visual analysis on informative rollout moments. Second, richer robot-state signals are converted into structured natural-language descriptions that can be jointly analyzed with visual observations. Together, these mechanisms combine the temporal precision of proprioception with the semantic reasoning capabilities of existing VLMs. When a small calibration set is available, \ourmethod{} can further leverage prediction errors as reflective experience in future prompts, providing lightweight adaptation through in-context learning and verbal reflection~\cite{brown2020language,kojima2022large,shinn2023reflexion}.

To enable systematic evaluation, we also introduce \emph{FailTime}, a robot manipulation benchmark with failure-onset annotations and synchronized visual and proprioceptive observations, combining re-annotated manipulation episodes with newly collected long-horizon real-world executions. Experiments show that modern VLMs already perform relatively well on binary failure detection under standard video prompting, while precise failure onset localization remains substantially harder. By integrating proprioception into VLM reasoning, \ourmethod{} enables sub-second failure onset localization in the zero-shot setting while also improving conventional diagnosis tasks. Ablations further show that proprioception is critical for temporal localization, while visual evidence remains essential for semantic understanding.

In summary, our contributions are three-fold: \textbf{1)} We introduce \emph{failure onset localization}, a new task for identifying the earliest failure-relevant deviation. \textbf{2)} We propose \ourmethod{}, a training-free framework that leverages proprioception for action-boundary localization and structured multimodal VLM prompting. \textbf{3)} We release \emph{FailTime}, a manipulation benchmark with failure-onset annotations and synchronized visual and proprioceptive observations.

%% file: icra/figures/teaser_title.tex
\begin{figure*}[!t]
  \centering
  \includegraphics[width=\textwidth]{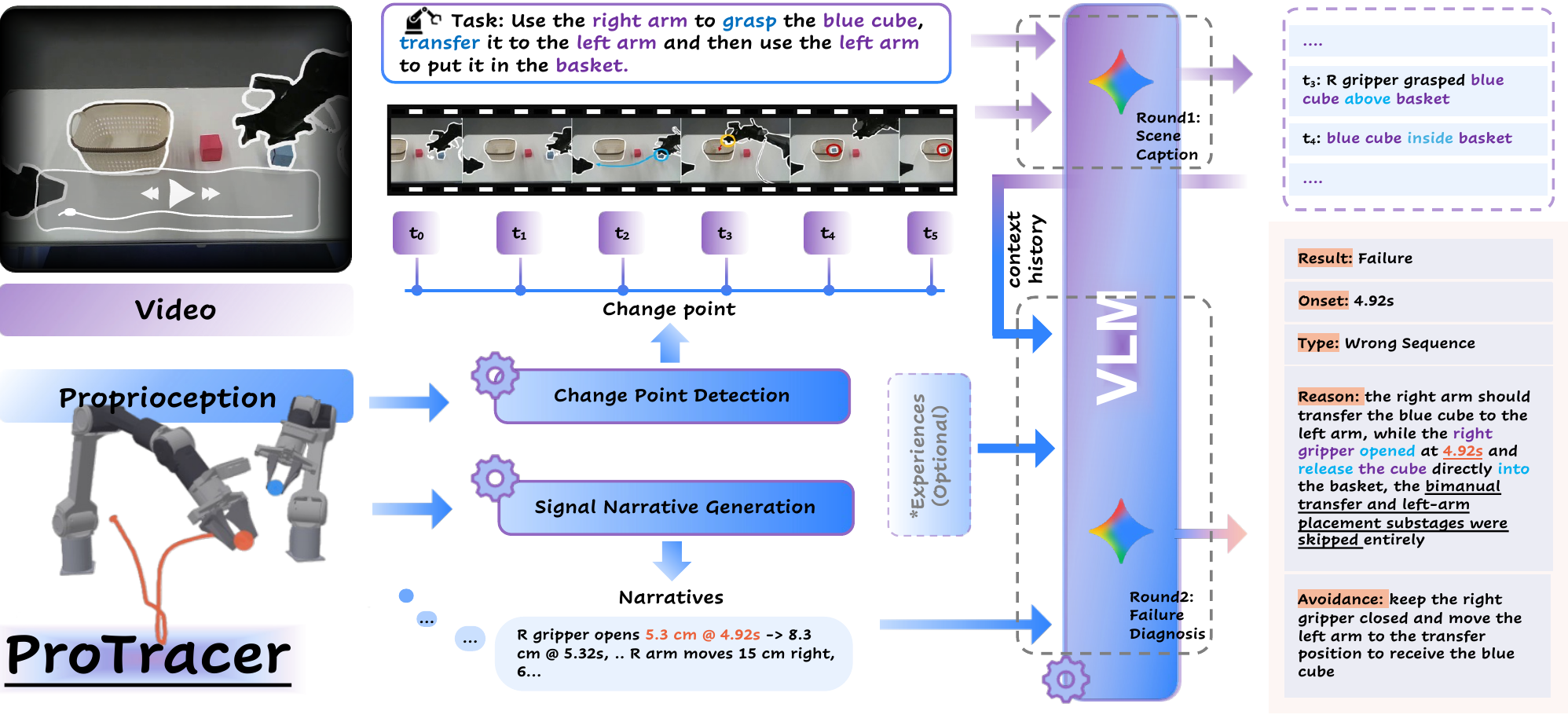}
  \caption{\textbf{\ourmethod{} Framework} for comprehensive robotic manipulation failure diagnosis, including failure detection, categorization, explanation, avoidance, and onset localization. It leverages proprioception for both keyframe selection and structured signal narratives generation. A VLM then jointly analyzes these multimodal inputs to infer the execution context and output the final diagnosis.}
  \label{fig:teaser}
\end{figure*}

%% file: icra/sections/related.tex
\section{Related Works}
\label{sec:related}

Our work relates to prior research on robot failure analysis, online failure detection, temporal localization,  proprioceptive reasoning, and VLM-based reasoning and adaptation.

\noindent\textbf{Robot Failure Analysis}
has been studied through failure explanation, communication, correction, and recovery~\cite{das2021explainable,dechant2023learning}. Recent works leverage foundation models and VLMs for failure diagnosis using multisensory summaries, video-QA supervision, visual symbols, or keyframe prompting~\cite{liu2023reflect,duan2024aha,robofac2025,vifailback,kite}, and for replanning~\cite{skreta2024replan}, recovery~\cite{raman2024cape}, and execution monitoring~\cite{zhou2024codeasmonitor}.These approaches primarily characterize whether and what went wrong, but not when the failure began.

\noindent\textbf{Online Failure Detection.}
Online error or anomaly detectors~\cite{agia2024unpacking,gu2025safe,romer2025failure,xu2025detect} operate causally on partial observations, where early failure evidence can be ambiguous and reliable detection may occur substantially after the actual onset. They also often require task- or policy-specific training data, nominal execution models, or threshold calibration. In contrast, we formulate onset localization as retrospective diagnosis, leveraging the complete rollout and outcome for more reliable temporal reasoning, which we empirically validate against online detectors in Sec.~\ref{sec:runtime}.

\noindent\textbf{Temporal Localization.}
Temporal video grounding and event spotting~\cite{hendricks2017localizing,grauman2022ego4d,deliege2021soccernetv2} localize visible events given text queries or predefined categories, whereas failure onset is the earliest deviation leading to eventual failure, conditioned on the task. Unlike VLM-based robot analysis that relies on dense or visually selected frames~\cite{duan2024aha,robofac2025,kite,vifailback}, we use proprioceptive dynamics to locate  action transitions.

\noindent\textbf{Proprioception-Guided Reasoning.}
Proprioceptive measurements such as gripper state, end-effector motion, and actuator dynamics are widely available in robot manipulation systems and datasets~\cite{james2020rlbench,mu2021maniskill,pumacay2024colosseum,khazatsky2024droid,padalkar2023open}, and have been used for control, manipulation learning, reward estimation, and trajectory generation~\cite{ma2022vip,mandlekar2023mimicgen,hoque2024intervengen,garrettskillgen}. We instead study their role in post-hoc VLM-based failure reasoning, using gripper and end-effector dynamics to identify action boundaries and converting richer robot-state signals into VLM-compatible descriptions.

\noindent\textbf{VLMs and Prompt-Level Adaptation.}
Our training-free framework builds on recent advances in VLM reasoning and prompt-level adaptation. VLMs have enabled language-conditioned planning, generalist policies, and visual reasoning in robotics~\cite{brohan2023can,brohan2023rt,liang2023code,huang2023voxposer,chen2024spatialvlm}. In-context learning enables pretrained models to adapt from examples without parameter updates~\cite{brown2020language}, while verbal reflection allows models to accumulate textual experience from previous errors~\cite{shinn2023reflexion,wang2023voyager}. Building on these ideas, we convert prediction errors on a small calibration set into textual lessons that guide subsequent failure diagnosis without VLM fine-tuning.

%% file: icra/sections/method.tex
\section{Problem Formulation \& Proposed Approach}
\label{sec:method}

\subsection{Diagnostic Setting}
\label{sec:problem_formulation}
We consider post-hoc failure diagnosis for robot manipulation episodes. Each episode contains a task instruction (e.g., ``Use the right arm to grasp the blue cube and place it into the basket''), one or more RGB video streams, and synchronized proprioceptive signals such as gripper states and end-effector poses. The objective is to determine whether the episode succeeds or fails and, for failed episodes, predict the failure type, generate an explanation and avoidance suggestion, and localize the failure onset moment.

Failure onset localization is the new diagnostic capability introduced in this work. Intuitively, the onset is the earliest moment at which the robot execution deviates from a valid task-completion trajectory in an episode that ultimately fails. We interpret an execution as a sequence of semantically meaningful action stages inferred from visual and proprioceptive evidence, such as reaching, grasping, lifting, transporting, and placing. Since the analysis is post-hoc and the full episode is available, temporary mistakes that are later corrected remain part of a valid execution; for example, a failed grasp followed by a successful retry is not treated as a failure onset. Thus, failure onset is defined only for episodes that ultimately fail, and corresponds to the earliest deviation after which the execution no longer returns to a successful task-completion trajectory.

This deviation can occur in three ways. First, the robot may transition away from an unsuccessful action without achieving its intended outcome, such as moving to placement without successfully grasping the object. Second, it may execute an action inconsistent with the task objective, such as placing the object in the wrong location or moving to an unrelated workspace region. Third, it may remain in the same action stage beyond an acceptable timeout without meaningful progress, such as repeatedly attempting to grasp without stable contact; in this case, the onset is the moment the timeout threshold is exceeded.

These rules provide an objective protocol for annotating failure onset without requiring exhaustive enumeration of all valid execution strategies or a complete task graph. Instead, the annotator only needs to judge whether the current execution remains consistent with \emph{some} valid task-completion trajectory. This is practical for post-hoc analysis: humans can typically identify when a robot is no longer ``on track'' without enumerating every possible valid execution.

\subsection{\ourmethod{} Overview}
Fig.~\ref{fig:teaser} depicts \ourmethod{}, a training-free VLM framework for post-hoc failure diagnosis that uses proprioception in two complementary ways: gripper and end-effector dynamics drive change-point detection to select informative transition moments and visual evidence, and robot-state measurements are converted into structured natural-language descriptions that the VLM analyzes jointly with the selected frames.

\subsubsection{Proprioceptive Analysis for Key Moment Detection\label{sec:keymoment}}
We use proprioceptive signals to identify candidate key moments in the execution. Joint and gripper states are converted into an 8D end-effector trajectory consisting of 3D position, quaternion orientation, and gripper aperture. For change-point detection, we focus on gripper dynamics and end-effector motion, which provide the most informative cues for manipulation transitions; the full 8D signal is used later for natural-language narrative generation.

Suppose the robot has $K$ arms. Let $v_t^k$ denote the gripper-aperture velocity of arm $k$ at time $t$, and let $m_t^k$ denote its denoised end-effector motion signal computed from consecutive positions. For each arm, we encode two discrete indicators: $\mathrm{sign}(v_t^k)\in\{-1,0,1\}$ and $\mathds{1}[m_t^k>0]\in\{0,1\}$. We discard magnitudes because primitive boundaries are primarily characterized by discrete changes in grasp state and motion activity rather than movement amplitude. For an episode of length $T$, this yields a symbolic matrix $X\in\mathbf{\Sigma}^{T\times 2K}$, where $x_t^c$ denotes channel $c$ at time $t$ and $\mathbf{\Sigma}$ is the induced discrete alphabet.

For a candidate segment $[s,e)$, we define the modal-Hamming cost as
{
\setlength{\abovedisplayskip}{2pt}
\setlength{\belowdisplayskip}{2pt}
\setlength{\abovedisplayshortskip}{2pt}
\setlength{\belowdisplayshortskip}{2pt}
\begin{equation}
    C(s,e) = \sum_{c=1}^{2K} \min_{v \in \Sigma_c} \sum_{t=s}^{e-1} \delta(x_t^c \neq v).
\end{equation}
} 
where $\delta(\cdot)$ is the indicator function and $\mathbf{\Sigma}_c$ is the alphabet of channel $c$. This cost measures how well a segment can be represented by a single stable symbolic state. Stable manipulation primitives yield low cost, while transitions between primitives produce higher inconsistency.

We apply PELT~\cite{pelt} to detect candidate change points, then greedily prune weak boundaries to satisfy a fixed temporal budget. For a boundary $b_i$ with neighboring boundaries $b_{i-1}$ and $b_{i+1}$, the removal cost increase is $S(b_i)=C(b_{i-1},b_{i+1})-C(b_{i-1},b_i)-C(b_i,b_{i+1})$. A small increase indicates that adjacent segments have similar proprioceptive states and can be merged with little loss. 
We iteratively remove the boundary with the smallest increase until at most $B$ temporal phases remain ($B{=}8$ on FailTime-Short and $39$ on FailTime-Long, with  $6.8$ and $19.5$ phases retained on average respectively). The retained boundaries define the selected keyframes and temporal intervals for downstream evidence construction.

\subsubsection{Proprioception-Guided VLM Inference}
\label{sec:vlm_inference}
Given the detected boundaries, \ourmethod{} extracts representative visual keyframes and converts proprioceptive signals into structured natural-language descriptions for joint reasoning with visual observations. Each interval is represented by keyframes, timestamps, and a rule-based sensor narrative summarizing the robot dynamics. These narratives describe gripper actions (opening, closing, holding), end-effector motion, and wrist rotation using displacement, motion direction, and rotation consistency, while merging short transient runs to reduce noise. They provide localized descriptions of robot behavior between sparse visual observations. We deliberately use deterministic rules rather than a learned captioner: the narration is hallucination-free, and its vocabulary consists of embodiment-agnostic physical quantities (aperture, displacement, rotation), which is why the same rules transfer unmodified to a different platform in Sec.~\ref{sec:experiments}.

VLM inference proceeds in two stages. First, the VLM analyzes selected keyframes from the available camera views, such as head-view and wrist-view images, and generates concise scene-state descriptions. To improve grounding consistency, we provide a task-specific reference vocabulary derived from the task instruction, covering relevant objects, actions, gripper states, gripper--object relations, and object--object relations. Second, the VLM receives the task instruction, visual keyframe descriptions, proprioceptive descriptions, failure taxonomy, and optional reflective experiences. The visual descriptions capture sparse scene states, while the proprioceptive descriptions characterize the physical transitions between them. From this combined prompt, the VLM constructs a concise event-level interpretation and predicts the final diagnosis.

\begin{figure}[!b]
  \centering
\includegraphics[width=1\columnwidth]{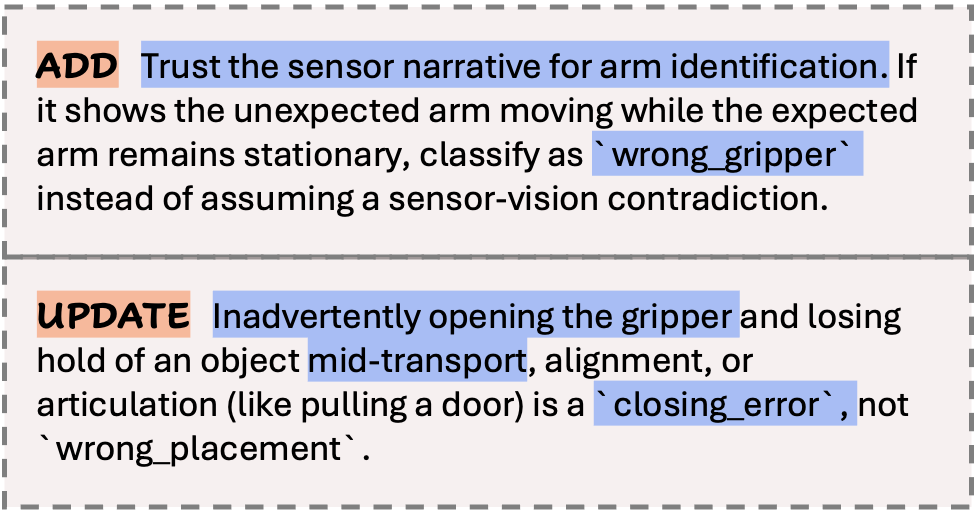}
\caption{\textbf{Examples of learned experiences.} }
  \label{fig:exp}
\end{figure}
\subsubsection{Optional Reflective Learning for Few-shot Adaptation}
\label{sec:reflective}
The preceding pipeline is training-free and can be applied directly to unseen robot embodiments and manipulation tasks.

When a small annotated calibration set is available for the target environment or task, \ourmethod{} can adapt to recurring error patterns that general-purpose VLM reasoning may miss without updating any weights.
To use this information without fine-tuning the VLM, we introduce an optional reflective memory for lightweight in-context adaptation.

Given a calibration set, the VLM first  predicts the failure verdict, onset time, and failure type for each episode. The prediction is then paired with the ground-truth labels and returned to the same VLM through a reflection prompt. The VLM analyzes the discrepancy and generates a  textual lesson, such as a common onset-localization error, a confusion between similar failure categories, or a mismatch between visual and proprioceptive evidence.

These lessons are accumulated into a  reflective experiences and prepended to future inference prompts. To keep the memory compact, the VLM may \emph{ADD} new lessons, \emph{MERGE} redundant observations, \emph{UPDATE} conflicting rules, or \emph{COMPRESS} the memory under a fixed word budget. Fig.~\ref{fig:exp} illustrates two real-world lessons successfully accepted into memory via in-context reflection. A candidate lesson is accepted only if re-running inference on the corresponding calibration episode with the updated memory corrects the original prediction; otherwise, it is discarded. This acceptance procedure is used only during adaptation. At test time, the memory is fixed and no ground-truth labels are used.

No VLM weights are updated, and no auxiliary reflector model is trained; adaptation occurs entirely through prompt-level contextual guidance. When no calibration data is available, the reflective memory is empty and \ourmethod{} reduces to the training-free proprioception-guided inference pipeline.

%% file: icra/sections/experiments.tex
\input{icra/tables/results_main}
\section{\ourdataset{} Dataset}
To evaluate failure onset localization, we introduce \emph{\ourdataset{}}, a robot manipulation dataset with frame-level onset annotations and detailed failure diagnoses, including 123 distinct manipulation tasks. It has two components. \emph{FailTime-Short} builds on the full ViFailback benchmark~\cite{vifailback} and a subset of its training data, yielding 1,539 annotated episodes, including 579 successful executions. \emph{FailTime-Long} is a newly collected set of 100 long-horizon real-world episodes (57 failures and 43 successes) on two bimanual platforms, a general-purpose ALOHA system and a cost-effective SO-101 setup. Its episodes average $54$\,s (max $120$\,s), much longer than the $12$\,s average (max $37$\,s) of FailTime-Short, so the onset must be located among far more candidate boundaries. It is designed to test cross-dataset transfer under different embodiments, scenes, and longer task horizons.

All episodes are annotated using the failure onset formulation in Sec.~\ref{sec:problem_formulation}. We develop an annotation interface for jointly inspecting synchronized video streams and aligned proprioceptive signals. Annotators are also given high-level task decompositions and example successful executions, while independently judging when the execution meaningfully deviates from a plausible successful trajectory. 

Beyond onset timestamps, we refine the original failure annotations by expanding the taxonomy from four to ten categories and correcting inconsistent reason and avoidance labels. The ten categories (wrong sequence, wrong object, wrong gripper, wrong placement, closing error, open error, translation error, orientation error, avoidable intervention, and unavoidable intervention), each failed episode receives one primary category, and each category has an explicit onset rule defining the earliest action that commits the robot to that failure. Six annotators with robotics and machine learning backgrounds spent over 120 hours annotating onset timestamps and failure categories. The resulting labels show strong consistency: average pairwise agreement is $0.94$ for failure categories, mean onset disagreement is $50$ms, and $99.6\%$ of onset disagreements fall within $0.5$s.

\section{Experiments}
\label{sec:experiments}
\subsection{Experimental Setup}

We perform a variety of experiments using \ourmethod{} with Gemini-3.1-pro as the underlying VLM backbone throughout. Performance is evaluated on the FailTime-Short and FailTime-Long benchmarks. For the optional reflective-adaptation variant, we construct the memory from 50 episodes randomly sampled from the unused portion of the training split, with no overlap with either the FailTime-Short or FailTime-Long evaluation set. We perform three reflective epochs. Sensitivity to both choices is analyzed in Table~\ref{tab:calib}. Since \ourmethod{} runs offline, it is not latency-constrained; for reference, an episode costs about \$0.09 and takes 40--60\,s on Gemini-3.1-pro, dominated by the two VLM calls.

\noindent\textbf{Baselines.}
We compare against several classes of baselines, most of which share the same Gemini-3.1-pro backbone as our method. First, we compare against Gemini-3.1-pro itself as a general-purpose VLM baseline, which uses standard dense prompting of multi-view video frames sampled at 1 fps. Second, we compare against Robotics-ER 1.6, a robotics-oriented model tuned by robotic data. Third, we compare against two methods specifically designed for robot failure diagnosis: KITE and ViFailback-8B. KITE is also based on Gemini-3.1-pro and uses optical-flow-based keyframe selection. For KITE, we preserve the original configuration whenever possible and modify only the prompting format to produce onset timestamps and the required diagnosis outputs. ViFailback-8B~\cite{vifailback} is a trained failure-analysis model originally developed for the ViFailback benchmark. Since it is trained on the original ViFailback training split, it is not a zero-shot baseline, unlike our approach and the other methods in the comparison. We additionally evaluate variants of these methods that receive proprioceptive signals as CSV files containing textual descriptions of feature types together with the corresponding numerical values. These variants are marked with an asterisk ($*$); ViFailback-8B, which is trained on the ViFailback training split, is marked with a dagger ($\dagger$).

\noindent\textbf{Metrics.}
For failure detection and failure-type classification, we report accuracy. Since some ambiguous episodes can contain multiple plausible failure types, a type prediction is considered correct if it matches any label in the ground-truth set. For failure onset grounding, we report tolerance-based accuracy~(\%)~\cite{deliege2021soccernetv2,hong2022spotting}, denoted $\mathrm{acc}@\delta$, where a prediction is correct if the absolute onset error is less than $\delta$. We use $\delta\in\{0.5s,1s\}$ and also report the mean absolute error (MAE) in seconds, computed only over correctly detected failure episodes. For failure reason and avoidance prediction, we use GPT-5 as an LLM judge following the rubric of ViFailback~\cite{vifailback}. For runtime detectors, we additionally report the area under the ROC curve (AUROC), together with the true-positive rate (TPR) and true-negative rate (TNR) at the calibrated operating threshold.

\input{icra/tables/sensor_vision_ablations}

\subsection{Main Results}

\noindent\textbf{Results on \ourdataset-Short.}
Table~\ref{tab:baseline_comparisons_table} shows that the primary challenge in \emph{\ourdataset{}-Short} is not binary failure detection, but temporally precise onset grounding. Dense-frame and failure-analysis baselines already achieve high failure detection accuracy, with all models exceeding $95\%$, whereas their onset accuracy varies widely. 
Access to proprioception already helps: simply appending raw proprioceptive signals to the same dense-frame Gemini-3.1-pro baseline (marked with * in Table~\ref{tab:baseline_comparisons_table}) reduces MAE from 1.32\,s to 0.90\,s and raises $acc@0.5s$ from $27.83\%$ to $48.03\%$, even without any change to the prompting scheme.
\ourmethod{} improves further, reaching $71.18\%$ $acc@0.5s$ and $0.59$\,s MAE zero-shot, and $77.34\%$ and $0.47$\,s with reflective experiences.
Proprioception is thus the source of the temporal signal, but how it is used matters as much: selecting key frames at proprioceptive change points and converting the signals into structured narratives further raises acc@0.5\,s from 48.03\% to 71.18\% and lowers MAE from 0.90\,s to 0.59\,s over raw sensor values alone, with Sec.~\ref{ablation} isolating each component.
Compared with ViFailback-8B, which is trained on the original ViFailback data, ProTracer still achieves substantially better strict temporal localization without any training. KITE, which likewise feeds a VLM with selected key frames but chooses them from visual cues, trails far behind ($14.53\%$ $acc@0.5s$); since replacing Sign-CPD with an optical-flow selector inside \ourmethod{} still yields $66.50\%$ (Table~\ref{tab:align}), the gap stems mainly from the proprioceptive narratives rather than frame selection alone. \ourmethod{} also improves failure-type classification to $77.83\%$, and gives the best failure reason and avoidance scores in Table~\ref{tab:baseline_comparisons_table} as well. 

\begin{figure}[!b]
  \centering
\includegraphics[width=\columnwidth]{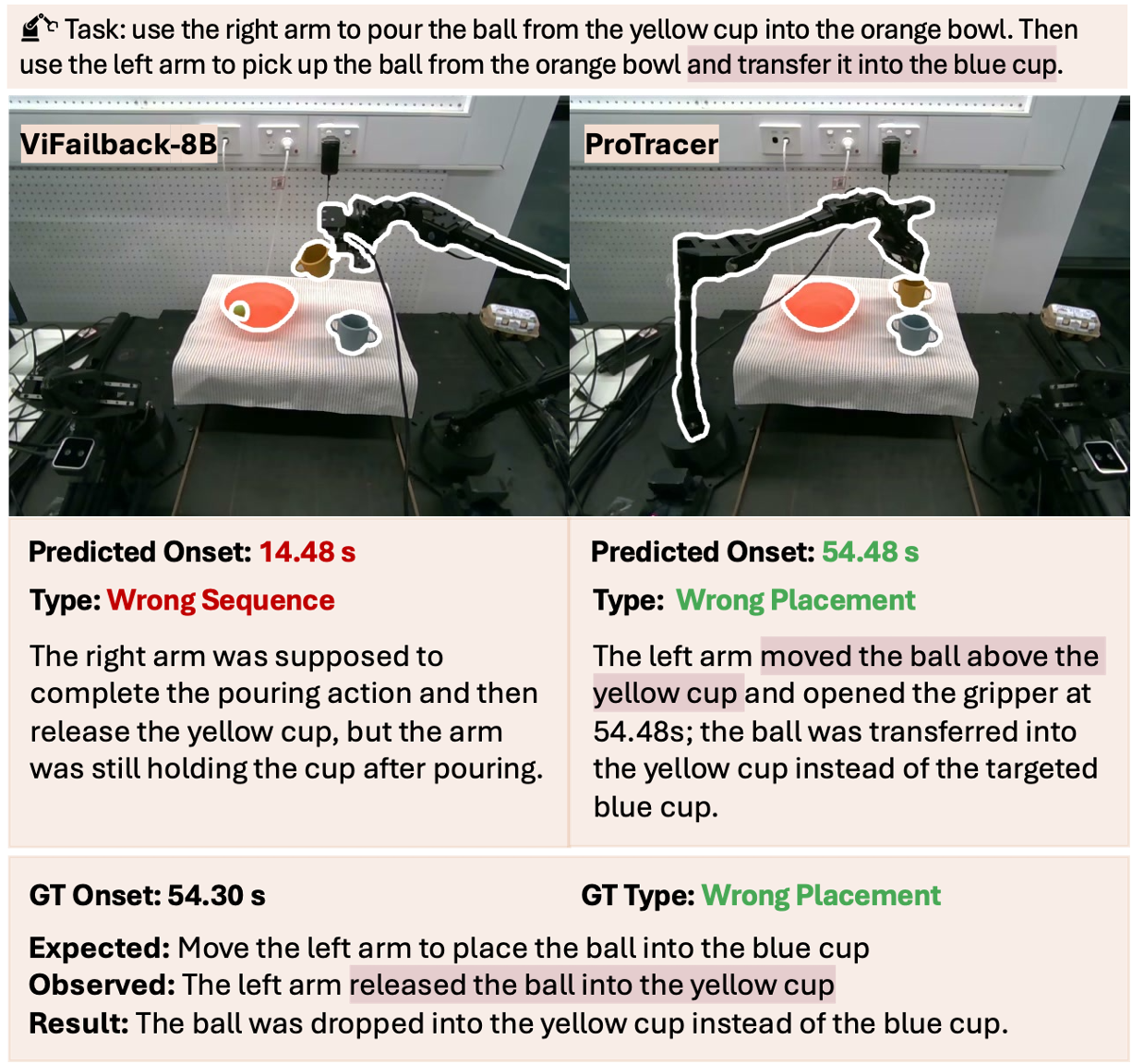}
  \caption{\textbf{Qualitative comparison on the \ourdataset-Long.} While ViFailback-8B incorrectly grounds the failure onset to an earlier stage, \ourmethod{} precisely localizes the true onset.}
  \label{fig:case_study}
\end{figure}
\noindent\textbf{Results on \ourdataset-Long.}
We further verify \ourmethod{} on long-horizon executions using \emph{\ourdataset{}-Long} in different embodiments and scenes (Table~\ref{tab:baseline_comparisons_table}). All methods degrade in onset localization on the longer horizons, but \ourmethod{} degrades the least, remaining the best on every onset and diagnosis metric while matching the best baseline on failure detection. ViFailback-8B, trained on ViFailback data and thus in-distribution for \emph{\ourdataset{}-Short}, drops from $46.06\%$ to $5.26\%$ at $acc@0.5s$, whereas training-free \ourmethod{} retains $40.35\%$ ($45.61\%$ with experiences). Fig.~\ref{fig:case_study} shows a qualitative comparison between ViFailback-8B and \ourmethod{}. Interestingly, Robotics-ER 1.6, whose training includes robot sensor data~\cite{geminirobotics2025}, degrades far less than Gemini-3.1-pro and is the strongest baseline on most metrics, again underscoring the value of proprioception, which \ourmethod{} supplies to an off-the-shelf VLM without training.

\subsection{Ablations and Analysis}
\label{ablation}
\begin{figure*}[!t]
  \centering
  \includegraphics[width=\textwidth]{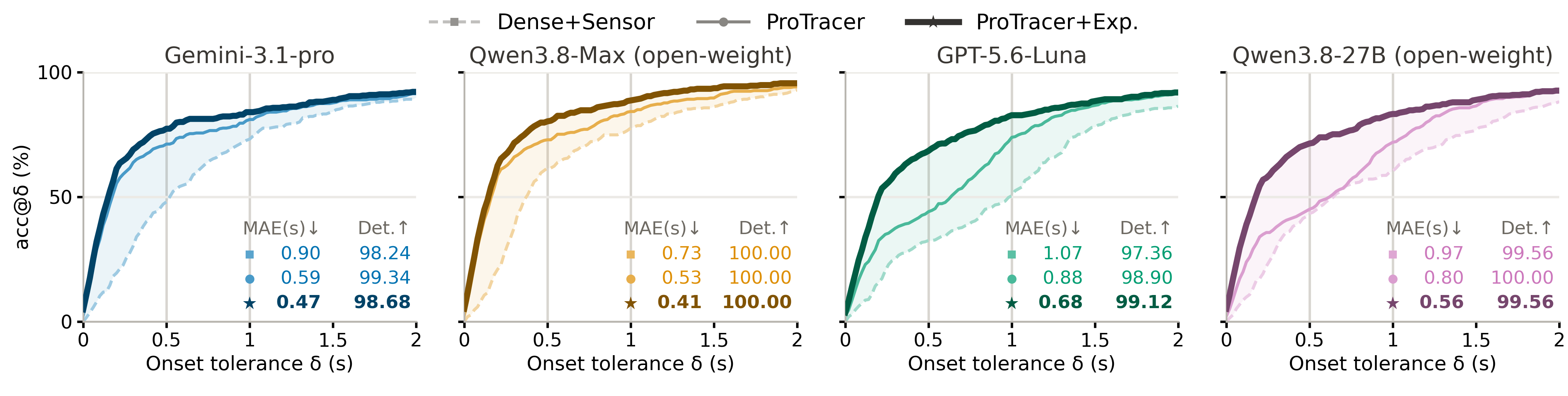}
\caption{\textbf{Generalization across VLM backbones on FailTime-Short.} Dense-frame prompting with raw proprioception vs.\ \ourmethod{} and \ourmethod{}+Exp; the reflective memory is learned with Gemini-3.1-pro and reused unchanged on the other backbones.}
\label{fig:backbones}
\end{figure*}
\input{icra/tables/calib_runtime}

\noindent\textbf{Modality Ablation.} Table~\ref{tab:ablation_sensor_vision} shows that proprioception and vision play complementary roles. Removing sensor input causes the largest drop in strict onset localization: $\mathrm{acc}@0.5s$ decreases from $71.18\%$ to $30.79\%$, and MAE doubles from $0.59$\,s to $1.18$\,s. This confirms that gripper and end-effector change points provide the temporal structure needed for sub-second grounding. Meanwhile, failure detection remains high ($98.24\%$) and type accuracy drops only moderately ($57.39\%$), so vision alone still supports semantic judgement. Removing vision instead collapses failure-type accuracy to $15.02\%$: without visual context, the model cannot determine what object-level failure occurred. These results support the design principle of \ourmethod{}: sensors determine \emph{when} to inspect, while vision-language reasoning determines \emph{what} failed.

\noindent\textbf{Key-frame Selection.}
\ourmethod{} selects key frames with Sign-CPD, a change-point detector over proprioceptive signals. Two questions follow: whether proprioception is truly necessary for this purpose or video-based cues such as optical flow could provide similar temporal information, and whether the symbolic sign-based formulation matters or a continuous-valued CPD on the same raw channels would do. Tab.~\ref{tab:align} answers both by swapping the key-frame selector inside the same pipeline while keeping everything else fixed. Sign-CPD is best on every diagnosis metric: onset accuracy within 0.5s improves from $66.50\%$ (optical flow) and $66.26\%$ (continuous CPD) to $71.18\%$, MAE from $0.73$s and $0.79$s to $0.59$s, and failure-type accuracy from $64.78\%$ and $66.50\%$ to $72.91\%$. To explain the gain, we ran a separate boundary-alignment experiment: we manually annotated action boundaries on 50 \emph{\ourdataset{}-Short} episodes with rich action transitions and measured how many of them each selector recovers within 0.5s. Sign-CPD recalls $69.68\%$, versus $40.80\%$ for optical flow and $55.28\%$ for continuous CPD, indicating that its key frames fall closer to the true primitive boundaries and suggesting that such boundaries are better captured by discrete state switches than by visual motion or amplitude changes.

\input{icra/sections/additional_analysis}

%% file: icra/tables/results_main.tex
\begin{table*}[!t]
\centering
\caption{\textbf{Results on FailTime-Short (AVG $12$\,s ,  MAX $37$\,s) and FailTime-Long (AVG $54$\,s , MAX $120$\,s ).} Variants marked $^*$ additionally receive raw proprioceptive signals; the one marked $^\dagger$ is trained on the ViFailback training split, hence not zero-shot.}
\label{tab:baseline_comparisons_table}
\renewcommand{\arraystretch}{1.0}
\resizebox{\textwidth}{!}{%
\begin{tabular}{l ccc cccc | ccc cccc}
\toprule
\multirow{3}{*}{\textbf{Model}}
& \multicolumn{7}{c|}{\textbf{FailTime-Short}}
& \multicolumn{7}{c}{\textbf{FailTime-Long}} \\
\cmidrule(lr){2-8} \cmidrule(lr){9-15}
& \multicolumn{3}{c}{\textbf{Failure Onset}}
& \multirow{2}{*}{\makecell{\textbf{Fail.}\\\textbf{Detect.}}}
& \multirow{2}{*}{\makecell{\textbf{Fail.}\\\textbf{Type}}}
& \multirow{2}{*}{\makecell{\textbf{Fail.}\\\textbf{Reason}}}
& \multirow{2}{*}{\makecell{\textbf{Fail.}\\\textbf{Avoid.}}}
& \multicolumn{3}{c}{\textbf{Failure Onset}}
& \multirow{2}{*}{\makecell{\textbf{Fail.}\\\textbf{Detect.}}}
& \multirow{2}{*}{\makecell{\textbf{Fail.}\\\textbf{Type}}}
& \multirow{2}{*}{\makecell{\textbf{Fail.}\\\textbf{Reason}}}
& \multirow{2}{*}{\makecell{\textbf{Fail.}\\\textbf{Avoid.}}} \\
\cmidrule(lr){2-4} \cmidrule(lr){9-11}
& MAE$\downarrow$ & $@0.5s$ & $@1s$ & & & &
& MAE$\downarrow$ & $@0.5s$ & $@1s$ & & & & \\
\midrule
\rowcolor{gray!15}
\textit{Failure Models} & & & & & & & & & & & & & & \\
ViFailback-8B$^\dagger$~\cite{vifailback} & 1.10 & 46.06 & 67.49 & 97.80 & 57.64 & 62.56 & 40.96 & 14.20 & 5.26 & 7.02 & 68.00 & 12.28 & 22.72 & 17.02 \\
KITE~\cite{kite} & 1.75 & 14.53 & 33.25 & 96.26 & 35.71 & 50.86 & 36.13 & 12.26 & 1.75 & 8.77 & 75.00 & 31.58 & 33.42 & 29.47 \\
\rowcolor{gray!15}
\textit{Robotic Models} & & & & & & & & & & & & & & \\
Robotics-ER 1.6~\cite{geminirobotics2025} & 1.63 & 14.29 & 36.21 & 95.38 & 34.48 & 49.06 & 24.06 & 6.50 & 14.91 & 34.21 & \textbf{88.00} & 37.72 & 38.54 & 34.78 \\
Robotics-ER 1.6$^*$~\cite{geminirobotics2025} & 1.42 & 33.50 & 44.83 & 96.70 & 35.71 & 48.89 & 26.88 & 5.82 & 17.54 & 30.70 & 83.00 & 40.35 & 38.84 & 26.45 \\
\rowcolor{gray!15}
\textit{General Models} & & & & & & & & & & & & & & \\
Gemini-3.1-pro & 1.32 & 27.83 & 55.42 & 95.82 & 49.51 & 54.30 & 41.26 & 10.22 & 8.77 & 21.05 & 67.00 & 38.60 & 35.53 & 31.23 \\
Gemini-3.1-pro$^*$ & 0.90 & 48.03 & 73.15 & 98.24 & 60.34 & 61.92 & 44.90 & 10.11 & 24.56 & 31.58 & 68.00 & 38.60 & 35.86 & 29.65 \\
\rowcolor{gray!15}
\textit{Ours} & & & & & & & & & & & & & & \\
\rowcolor{ourshl} \textbf{\ourmethod{}} & 0.59 & 71.18 & 81.03 & \textbf{99.34} & 72.91 & \textbf{69.40} & 50.87 & 4.06 & 40.35 & 49.12 & 87.00 & 57.89 & 49.19 & 40.09 \\
\rowcolor{ourshl} \textbf{\ourmethod{}+Exp.} & \textbf{0.47} & \textbf{77.34} & \textbf{83.99} & 98.68 & \textbf{77.83} & 69.20 & \textbf{51.24} & \textbf{3.27} & \textbf{45.61} & \textbf{50.88} & \textbf{88.00} & \textbf{59.65} & \textbf{50.16} & \textbf{40.18} \\
\bottomrule
\end{tabular}%
}
\end{table*}

%% file: icra/tables/sensor_vision_ablations.tex
\begin{table*}[!t]
\centering
\begin{minipage}[t]{0.48\textwidth}
\centering
\caption{\textbf{Modality ablations on FailTime-Short.}}
\label{tab:ablation_sensor_vision}
\renewcommand{\arraystretch}{0.9}
\setlength{\tabcolsep}{7pt}
\scriptsize
\begin{tabular}{l ccc cc}
\toprule
\multirow{2}{*}{\textbf{Model}}
& \multicolumn{3}{c}{\textbf{Failure Onset}}
& \multirow{2}{*}{\makecell{\textbf{Fail.}\\\textbf{Detect.}}}
& \multirow{2}{*}{\makecell{\textbf{Fail.}\\\textbf{Type}}} \\
\cmidrule(lr){2-4}
& MAE$\downarrow$ & $@0.5s$ & $@1s$ & & \\
\midrule
\rowcolor{ourshl} \textbf{\ourmethod{}} (Sign-CPD) & \textbf{0.59} & \textbf{71.18} & \textbf{81.03} & \textbf{99.34} & \textbf{72.91} \\
\hspace{1em} w/o sensor & 1.18 & 30.79 & 59.85 & 98.24 & 57.39 \\
\hspace{1em} w/o vision & 1.88 & 23.65 & 48.52 & 90.33 & 15.02 \\
\bottomrule
\end{tabular}
\end{minipage}
\hfill
\begin{minipage}[t]{0.48\textwidth}
\centering
\renewcommand{\arraystretch}{0.9}
\scriptsize
\caption{\textbf{Key-frame selectors on FailTime-Short.}}
\label{tab:align}
\setlength{\tabcolsep}{5pt}
\begin{tabular}{l ccc cc}
\toprule
\multirow{2}{*}{\textbf{Selector}} & \multicolumn{3}{c}{\textbf{Failure Onset}} & \multirow{2}{*}{\makecell{\textbf{Fail.}\\\textbf{Detect.}}} & \multirow{2}{*}{\makecell{\textbf{Fail.}\\\textbf{Type}}} \\
\cmidrule(lr){2-4}
 & MAE$\downarrow$ & $@0.5s$ & $@1s$ & & \\
\midrule
\textbf{ProTracer }(Optical Flow) & 0.73 & 66.50 & 76.11 & 98.68 & 64.78 \\
\textbf{ProTracer} (Continuous CPD) & 0.79 & 66.26 & 77.34 & 98.90 & 66.50 \\
\rowcolor{ourshl} \textbf{ProTracer} (\textbf{Sign-CPD}) & \textbf{0.59} & \textbf{71.18} & \textbf{81.03} & \textbf{99.34} & \textbf{72.91} \\
\bottomrule
\end{tabular}
\end{minipage}
\end{table*}

%% file: icra/tables/calib_runtime.tex
\begin{table*}[!t]
\centering
\begin{minipage}[t]{0.48\textwidth}
\centering
\caption{\textbf{Calibration-set size and reflective epochs.}}
\label{tab:calib}
\scriptsize
\setlength{\tabcolsep}{4pt}
\renewcommand{\arraystretch}{0.95}
\begin{tabularx}{\linewidth}{@{}l*{5}{>{\centering\arraybackslash}X}@{}}
\toprule
Setting & MAE$\downarrow$ & @0.5s$\uparrow$ & @1s$\uparrow$ & Det. $\uparrow$ & Type $\uparrow$ \\
\midrule
ProTracer \emph{(w/o Exp.)} & 0.59 & 71.18 & 81.03 & \textbf{99.34} & 72.91 \\
\midrule
\multicolumn{6}{@{}l}{\emph{Calibration-set size}} \\
25 eps  & 0.55 & 73.65 & 81.28 & 98.68 & 76.85 \\
\rowcolor{ourshl} 50 eps  & \textbf{0.47} & \textbf{77.34} & \textbf{83.99} & 98.68 & \textbf{77.83} \\
100 eps & \textbf{0.47} & 75.62 & 82.51 & 98.68 & 76.60 \\
\midrule
\multicolumn{6}{@{}l}{\emph{Reflective epochs (50 eps)}} \\
1 epoch  & 0.56 & 74.38 & 81.77 & 98.68 & 71.18 \\
2 epochs & 0.52 & 74.14 & 83.74 & 98.90 & 77.09 \\
\rowcolor{ourshl} 3 epochs & \textbf{0.47} & \textbf{77.34} & \textbf{83.99} & 98.68 & \textbf{77.83} \\
\bottomrule
\end{tabularx}
\end{minipage}
\hfill
\begin{minipage}[t]{0.48\textwidth}
\centering
\caption{\textbf{Runtime failure detectors on FailTime-Short.} }
\label{tab:runtime}
\scriptsize
\setlength{\tabcolsep}{4pt}
\renewcommand{\arraystretch}{0.95}
\begin{tabularx}{\linewidth}{@{}l*{6}{>{\centering\arraybackslash}X}@{}}
\toprule
\multirow{2}{*}{\textbf{Detector}} & \multicolumn{3}{c}{\textbf{Failure Detection}} & \multicolumn{3}{c}{\textbf{Failure Onset}} \\
\cmidrule(lr){2-4} \cmidrule(lr){5-7}
 & AUROC & TPR & TNR & MAE$\downarrow$ & @0.5s & @1s \\
\midrule
RND-OE~\cite{xu2025detect}      & 0.57 & 44.83 & 79.59  & 5.03 & 0.74  & 2.96  \\
FIDeL~\cite{fidel2026}              & 0.60 & 34.73 & 95.92  & 5.27 & 0.25  & 1.97  \\
RynnValue-8B~\cite{rynnvalue2026}   & 0.85 & 38.18 & 93.88  & 2.61 & 1.72  & 5.67  \\
GRU (Onset Supervised)                       & 0.74 & 71.43 & 65.31  & 2.08 & 18.97 & 33.99 \\
\midrule
\rowcolor{ourshl} \textbf{\ourmethod{}} (zero-shot)      & --   & \textbf{99.26} & \textbf{100.00} & \textbf{0.59} & \textbf{71.18} & \textbf{81.03} \\
\bottomrule
\end{tabularx}
\end{minipage}
\end{table*}

%% file: icra/sections/additional_analysis.tex


\myheading{Generalization across VLM backbones.}
To verify that the onset-localization gains are not tied to \textbf{Gemini-3.1-pro}, Fig.~\ref{fig:backbones} repeats the comparison between dense-frame prompting with raw proprioception (Dense+sensor), \ourmethod{}, and \ourmethod{}+Exp with three additional backbones: \textbf{Qwen3.8-Max}, \textbf{GPT-5.6-Luna}, and \textbf{Qwen3.8-27B}. \ourmethod{} improves over Dense+sensor on the three additional backbones for onset localization with lower MAE, confirming that the benefit comes from structuring proprioception into boundary evidence and narratives rather than from access to raw sensor values. The consistent separation of the curves across backbones further shows that the improvement holds for models of different capacities.

\noindent\textbf{Sensitivity to calibration-set size and reflective epochs.}
Table~IV varies the calibration-set size and the number of reflective epochs. Every configuration improves onset performance over ProTracer without experiences, so the benefit of reflective memory does not depend on a particular setting. Using more than 50 calibration episodes brings no further gain, suggesting that a small calibration set suffices.

\myheading{Reflective experience transfer.}
The reflective experience transfers in two ways. First, a memory built on \emph{\ourdataset{}-Short} also improves \emph{\ourdataset{}-Long} (Table~\ref{tab:baseline_comparisons_table}). Second, the same memory, learned with Gemini-3.1-pro and reused unchanged, consistently improves onset localization, with particularly large gains for the weaker models (Fig.~\ref{fig:backbones}). This suggests that reflective experience captures transferable knowledge rather than model-specific prompt tuning.



\subsection{Comparison with Runtime Failure Detectors}
\label{sec:runtime}
\begin{figure}[!t]
  \centering
  \vspace{-8pt}   
  \includegraphics[width=1\columnwidth]{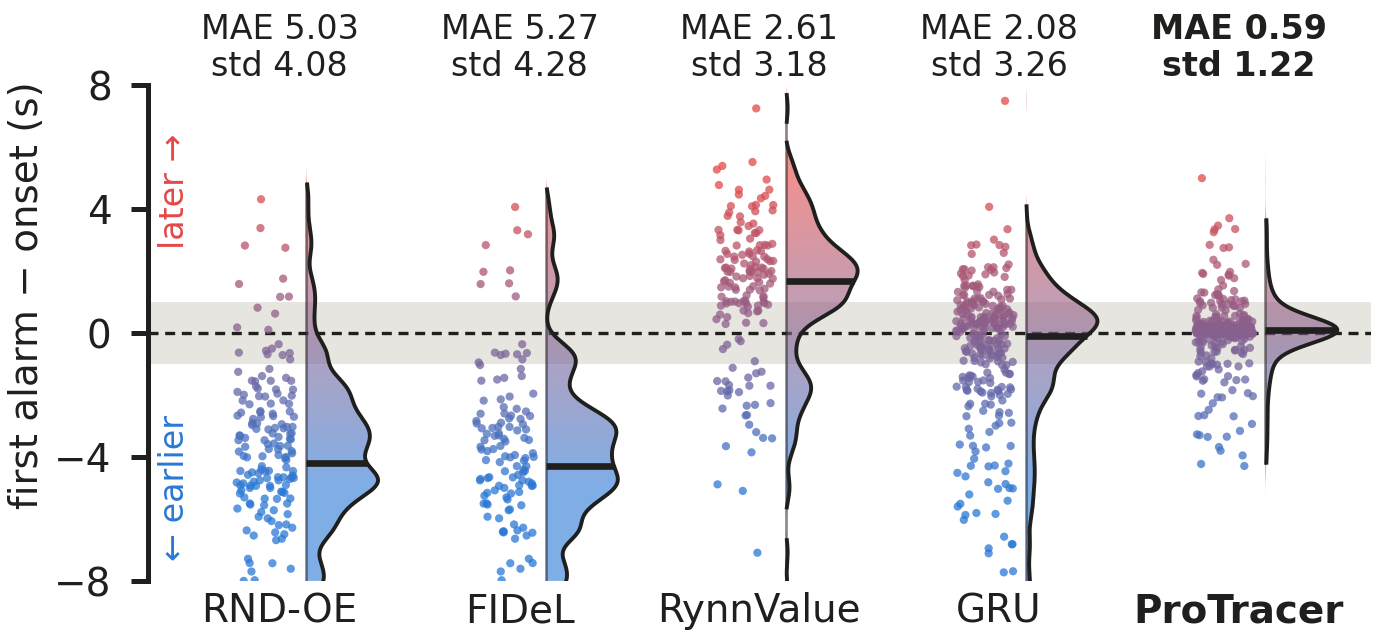}
  \caption{\textbf{First-alarm error relative to the annotated onset on FailTime-Short} (alarmed episodes only, clipped to $\pm 8$\,s; band: $1$\,s tolerance).}
  \label{fig:runtime_dist}
\end{figure}
Runtime failure detectors are designed to stop a policy, not to localize where an execution went wrong, but a natural question is whether their first alarm could serve as an onset estimate. Table~\ref{tab:runtime} tests this on FailTime-Short by running each detector causally over every episode and taking its first alarm as the onset estimate; episodes without an alarm count as misses. We include the \textbf{RND-OE} uncertainty score \cite{xu2025detect}, \textbf{FIDeL}~\cite{fidel2026} (statistical filter), a value-drop detector on a finetuned \textbf{RynnValue-8B}~\cite{rynnvalue2026}, and, as a strong reference, a \textbf{GRU} trained with onset labels on the FailTime-Short evaluation episodes themselves via 5-fold cross-validation, giving it in-distribution supervision that others do not use. All detectors are run policy-agnostically on frozen visual features and, following the original methods, thresholded to a $5\%$ false-alarm rate on held-out successful demonstrations. Runtime alarms are poor onset estimates. The three label-free detectors, which are fit on successful rollouts only, reach at most $1.7\%$ $acc@0.5s$ with MAEs of 2.6--5.3\,s, and even the supervised GRU only $19.0\%$ and 2.08\,s, against $71.2\%$ and 0.59\,s for \ourmethod{} without any training. 
Fig.~\ref{fig:runtime_dist} shows why the gap cannot be calibrated away: the two density-based detectors fire systematically early and the value model late, but with a spread of 3--4\,s around their medians, so shifting every alarm of a detector by its own median offset, an oracle correction estimated on the test set, still leaves onset accuracy below $9\%$ $acc@0.5s$ ($18.7\%$ for the GRU).
Runtime detection and onset localization thus answer different questions. Runtime detectors decide when to stop under causal constraints, and doing so reliably requires nominal data for calibration and a tolerance for early alarms; \ourmethod{} instead recovers where an execution went wrong from the full recording without any training, which makes it the better fit for post-hoc analysis and root-cause tracing.

%% file: icra/sections/conclusion.tex
\section{Conclusion}
\label{sec:conclusion}

In this paper, we introduced failure onset localization as a new capability for robot manipulation failure diagnosis, together with FailTime, a benchmark with synchronized visual and proprioceptive observations for evaluating fine-grained temporal failure analysis. We further proposed ProTracer, a training-free framework that integrates visual and proprioceptive signals for multimodal failure reasoning using existing VLMs. Experimental results demonstrate strong performance across both conventional failure diagnosis tasks and the newly introduced onset localization task, highlighting the importance of proprioceptive reasoning for precise robot failure understanding.

%% file: icra/appendix/analysis.tex
\section{Additional Experimental Analyses}
\label{app:analysis}

\myheading{Inference cost.}
As an offline post-hoc method, \ourmethod{} is not latency-constrained, unlike runtime failure detectors. For reference, on Gemini-3.1-pro an episode costs approximately \$0.09 (batched) and takes 40--60\,s on average, dominated by the two VLM calls.

\input{icra/appendix/inputs/reflection_learning_table}

Table~\ref{tab:reflection_learning_tables} provides the per-iteration breakdown behind the epoch sweep in Table~\ref{tab:calib}, showing that accumulating experience improves not only calibration (train) performance but also test performance, confirming that the learned rules generalize rather than overfit. Table~\ref{tab:qwen27b} further shows that experience learning transfers across models: although Qwen3-VL-27B-think performs poorly overall due to its small size, experience still improves it on every metric, mirroring the trend on our main backbone.

%% file: icra/appendix/inputs/reflection_learning_table.tex
\begin{table*}[!t]
\centering
\caption{\textbf{Effect of experience accumulation in FailTime-Short.}}
\label{tab:reflection_learning_tables}
\renewcommand{\arraystretch}{1.0}
\setlength{\tabcolsep}{4pt}
\scriptsize
\begin{tabular}{l|ccccc|ccccc}
\toprule
& \multicolumn{5}{c|}{\textbf{train}}
& \multicolumn{5}{c}{\textbf{test}} \\
\cmidrule(lr){2-6}\cmidrule(lr){7-11}
\textbf{Epochs}
 & \multicolumn{3}{c}{\textbf{Failure Onset}} & \textbf{Failure} & \textbf{Failure}
 & \multicolumn{3}{c}{\textbf{Failure Onset}} & \textbf{Failure} & \textbf{Failure} \\
\cmidrule(lr){2-4}\cmidrule(lr){7-9}
 & MAE$\downarrow$ & $\delta{=}0.5s$ & $\delta{=}1s$ & \textbf{Detection} & \textbf{Type}
 & MAE$\downarrow$ & $\delta{=}0.5s$ & $\delta{=}1s$ & \textbf{Detection} & \textbf{Type} \\
\midrule
0 \emph{(w/o Exp.)} & 0.41 & 72.50 & 82.50 & 98.00 & 75.00
                    & 0.59 & 71.18 & 81.03 & \textbf{99.34} & 72.91 \\
1   & 0.48 & 70.00 & 87.50 & 98.00 & 70.00
     & 0.56 & 74.38 & 81.77 & 98.68 & 71.18 \\
2  & \textbf{0.33} & 70.00 & 85.00 & 96.00 & 72.50
     & 0.52 & 74.14 & 83.74 & 98.90 & 77.09 \\
3  & \textbf{0.33} & \textbf{77.50} & \textbf{90.00} & 98.00 & 75.00
     & \textbf{0.47} & \textbf{77.34} & \textbf{83.99} & 98.68 & \textbf{77.83} \\
\bottomrule
\end{tabular}
\vspace{1.2em}

\centering
\scriptsize
\setlength{\tabcolsep}{4pt}
\caption{\textbf{Qwen3.6-27B results on FailTime-Short.}.}
\label{tab:qwen27b}
\begin{tabular}{l ccc cc cc}
\toprule
\multirow{2}{*}{Model}
 & \multicolumn{3}{c}{\textbf{Failure Onset}}
 & \textbf{Failure} & \textbf{Failure} & \textbf{Failure} & \textbf{Failure} \\
\cmidrule(lr){2-4}
 & MAE(s)$\downarrow$ & $\delta{=}0.5$\,s & $\delta{=}1$\,s
 & \textbf{Detection} & \textbf{Type} & \textbf{Reason} & \textbf{Avoidance} \\
\midrule
\textbf{Qwen3-VL-27B-think (zero-shot) }      & 1.63 & 26.60 & 46.55
                                      & 94.51 & 26.60 & 36.19 & 24.17 \\
\textbf{Qwen3-VL-27B-think (+ EXP.) }& \textbf{1.50} & \textbf{36.45} & \textbf{53.20}
                                      & \textbf{95.38} & \textbf{36.70} & \textbf{41.50} & \textbf{27.16} \\
\bottomrule
\end{tabular}
\end{table*}

%% file: icra/appendix/taxonomy.tex
\input{icra/appendix/inputs/taxonomy_full}
\section{Failure Taxonomy and Annotation Fields}
\label{app:taxonomy}

Each failed episode is annotated along five fields: \emph{failure detection}, \emph{failure onset}, \emph{failure type}
\emph{failure reason}, and \emph{failure avoidance}.

\textbf{Failure Detection.} The verdict is a binary classification: an episode
is labeled  \emph{Success} if the robot completed all required steps in the specified order according to the task description; \emph{failure} if the robot did not complete the task or violated the required order according to the task description.

\textbf{Failure Onset.} The onset is the earliest moment at which the robot execution deviates from a valid task-completion trajectory in an episode that ultimately fails. We operationalize this as the first commit action that lies outside the current substage's expected action and locks in the failure (i.e.\ the earliest deviation that is not subsequently recovered and is followed by task failure.). Approach, alignment, and other recoverable repositioning are not onsets: a deviation the robot later recovers from is not a failure onset. For example, in a pick-up decomposed as move\,$\rightarrow$\,grasp\,$\rightarrow$\,move, if the object is not secured, the onset is the moment the robot starts the next substage(lifting or moving away) while the object is not held; an initial failed grasp that is then successfully re-grasped has no onset. If a substage stalls with no clear commit action, we fall back to $\min(\text{next\_substage\_start},\ \text{video\_end})$, and cascading failures share the root onset. specialized per failure mode in Table~\ref{tab:taxonomy_full}. 

\textbf{Failure Reason.} The failure reason captures the concrete mismatch between expected and observed outcome, in three parts: what the plan expected, what was observed, and the resulting failure. It is stated in
physical terms(object relationships, aperture state, pose offset, target identity, final object state). No speculation about why the policy chose its action.

\textbf{Failure Type.} We assign each failed episode a single \emph{primary}
failure type from a taxonomy of ten classes (Table~\ref{tab:taxonomy_full}).  A single failure can legitimately fall under several types (e.g.\ a \texttt{translation\_error} that leads to a \texttt{wrong\_placement}), so a prediction is counted as correct if it matches any of the labeled types.

\textbf{Failure Avoidance.} The smallest concrete physical action before onset that would close the expected $-$ observed gap. Not meta-advice and not a re-plan.

%% file: icra/appendix/inputs/taxonomy_full.tex
\begin{table*}[t]
  \centering
  \scriptsize
\newlength{\ttype}\setlength{\ttype}{3.2cm}
\newlength{\trest}%
\setlength{\trest}{\dimexpr(\textwidth-\ttype-4\tabcolsep)/2\relax}
  \renewcommand{\arraystretch}{1.25}
  \caption{\textbf{The failure taxonomy. }Each failed episode is assigned one primary
  type.}
\begin{tabular}{@{}
    >{\raggedright\arraybackslash}p{\ttype}
    >{\raggedright\arraybackslash}p{\trest}
    >{\raggedright\arraybackslash}p{\trest}@{}}
    \toprule
    \textbf{Type} & \textbf{Definition} & \textbf{Onset rule} \\
    \midrule
    \texttt{wrong\_sequence} & Substages performed in the wrong order, or a
    required substage skipped entirely. & Wrong-substage contact/commit start,
    or wrong-substage end if no task-relevant contact occurs. \\

    \texttt{wrong\_object} & The robot commits to the wrong target object; a
    wrong-object approach alone is not onset if it can still retarget. &
    Wrong-object contact/close/release start, or the wrong targeted substage end
    if no task-relevant contact occurs. \\

    \texttt{wrong\_gripper} & The wrong arm performs a substage or atomic action
    meant for the other arm. & Wrong-gripper contact start, or wrong-gripper
    substage end if no task-relevant contact occurs. \\

    \texttt{wrong\_placement} & A firmly-held object is released at the wrong
    target location. & Release start at the wrong target location. \\

    \texttt{closing\_error} & The correct gripper attempts to grasp or keep
    grasping the correct object but fails to secure or retain hold of it. &
    Next-substage start: the robot lifts, moves away, or transports while the
    object is not securely held. \\

    \texttt{open\_error} & The gripper fails to open when required: it does not
    release a held object, or does not pre-open wide enough before grasping. &
    Contact or release start, when the gripper is visibly not open enough:
    contact without sufficient pre-opening, or reaching release while still
    closed. \\

    \texttt{translation\_\allowbreak error} & The gripper's XYZ position is
    misaligned with the object/target while wrist orientation is acceptable; a
    position shift alone would fix the action. & Contact/close/release start
    with the offset pose. \\

    \texttt{orientation\_\allowbreak error} & The gripper's yaw/pitch/roll is
    misaligned with the object's affordance orientation (e.g.\ opening axis
    along the grasp axis), so correct XYZ alone would not rescue the grasp. &
    Contact/close/release start with the wrong orientation. \\

    \texttt{avoidable\_\allowbreak intervention} & An external disturbance
    (typically a human moving an object mid-episode) triggered a downstream
    failure the robot could have avoided by retargeting. & The underlying
    failure's onset, under a disturbed but still recoverable scene. \\

    \texttt{unavoidable\_\allowbreak intervention} & An external disturbance
    left the scene unrecoverable (target removed or blocked); the arm proceeded
    with no way to rescue. & Commit-action start under the disturbed scene, when
    the target or required state is no longer physically available. \\
    \bottomrule
  \end{tabular}
  \label{tab:taxonomy_full}
\end{table*}

%% file: icra/appendix/dataset.tex
\section{Details of FailTime}
\label{sec:dataset}

\subsection{FailTime-Short and FailTime-Long}

The tasks in FailTime-Short are drawn from the ViFailback dataset~\cite{vifailback}, which spans 109 manipulation tasks. We annotated over 1{,}500 episodes in total, of which we adopt ViFailback's official test split of $455$ episodes with median of $9.2$\,s, (range $0.7$--$30.9$\,s) for all main experiments. In contrast, our newly collected FailTime-Long targets long-horizon multi-stage manipulation: it consists of $30$ episodes drawn from the 8 tasks listed in Tab.~\ref{tab:failtime-alh-tasks}, with a median length of $67.4$\,s (range $22$--$116$\,s), which is roughly $7\times$ longer per episode than FailTime-Short and far more demanding for temporal reasoning over extended action sequences.

\input{icra/appendix/inputs/Task_description_ALH}
\subsection{Annotation Interface}
We design an annotation interface to streamline the labeling process for annotators, as shown in Fig.~\ref{fig:dataset_annotation}. For each episode, an annotator selects one failed task substage which either inherited from the original ViFailback dataset or decomposed by an LLM, and optionally edited by the annotator to align with our taxonomy, and identifies the onset time based on both the proprioceptive signals and the multi-view videos, then provides the failure type, reason, and avoidance strategy. Annotators are required to follow our predefined taxonomy and to describe the \emph{Observed} behavior and the \emph{Result}; the \emph{Expected} behavior is derived directly from the task substage definitions and therefore does not need to be annotated.
\input{icra/appendix/inputs/annotation}

%% file: icra/appendix/inputs/Task_description_ALH.tex
\begin{table*}[t]
\centering
\scriptsize
\setlength{\tabcolsep}{6pt}
\caption{\textbf{FailTime-Long task catalogue.} The benchmark consists of 30
challenging long-horizon episodes drawn from 8 distinct tasks executed on
an ALOHA bimanual platform. All recorded trajectories are released for
the benchmark evaluation.}
\label{tab:failtime-alh-tasks}
\renewcommand{\arraystretch}{1.25}
\begin{tabularx}{\textwidth}{c X}
\toprule
\textbf{Task ID} & \textbf{Task Description}\\
\midrule
\texttt{task\_1} & Use the right arm to pour the ball from the yellow cup into the orange bowl. Then use the left arm to pick up the ball from the orange bowl and transfer it into the blue cup. \\
\texttt{task\_2} & Use the right arm to pick up all blocks from the bowl and place them on the table. Then use the left arm to assemble the small block into the second slot from the right of the 4 slot block. \\
\texttt{task\_3} & Place all balls into the six slots egg carton: Use the right arm to place the three balls on the right side of the workspace, and use the left arm to place the balls on the left side of the workspace. \\
\texttt{task\_4} & Use the right arm to separate the six nested upside-down paper cups and stack them into a 3-2-1 pyramid structure. \\
\texttt{task\_5} & Use the right arm to remove the cup from the bottle then transfer the bottle to the bowl. Then use the left arm to place the white object to mask over the bottle.\\
\texttt{task\_6} & Use the right arm to pick up the marker and align it to the left arm to grasp the cap. Use the left arm to hold it, and use the right arm to remove the cap. Then use the right arm to write a line on the whiteboard. Finally, use the right arm to align the marker with the cap held by the left arm, to cap the marker, and place it into the cup.\\
\texttt{task\_7} & Use the right arm to place the yellow block into the yellow bowl on the left, then use the left arm to place the red block into the orange ball on the right.\\
\texttt{task\_8} & Use the left arm to contact and stabilize the green block, then use the right arm to assemble the small block into its rightmost slot. \\
\bottomrule
\end{tabularx}
\end{table*}

%% file: icra/appendix/inputs/annotation.tex
\begin{figure}[h]
    \centering
    \includegraphics[width=\linewidth]{figures/annotation tool.png}
    \caption{The annotation interface of FailTime Dataset}
    \label{fig:dataset_annotation}
\end{figure}

%% file: icra/appendix/method.tex
\section{Details of \ourmethod{}}
\label{sec:app_method}

\subsection{Proprioceptive Analysis for Key Moment Detection}
To illustrate how Sign-CPD selects keyframes, Fig.~\ref{fig:cpd_demo} visualizes detected boundaries and selected frames on two representative episodes. In the left pick-and-place example, the right arm executes a sequence of primitives: approach-align $\rightarrow$ grasp $\rightarrow$ transport $\rightarrow$ release; Sign-CPD segments the transitions between these actions and selects salient frames around the corresponding phase boundaries. The right example shows a bimanual pick-and-place task, where Sign-CPD similarly recovers meaningful boundaries across both arms.

We further compare Sign-CPD with an optical-flow baseline using human-annotated primitive boundaries on 50 successful episodes. As shown in Fig.~\ref{fig:primitive_alignment}, proprioceptive change points align more closely with annotated primitive transitions than visual motion peaks, supporting our use of robot-state dynamics as temporal priors for VLM reasoning.

\input{icra/appendix/inputs/cpd_demo}

\input{icra/appendix/inputs/primitive_alignment}

\subsection{Proprioceptive Signal Description}
\label{sec:Proprioceptive Signal Description}
The proprioceptive data consists of informative physical signals about how each arm moves and interacts with objects, but the raw stream is too dense for an VLM to digest directly. For every frame and every arm the robot exposes an $8$-dimensional state vector $(x,y,z,q_x,q_y,q_z,q_w,g)$ , where three for the end-effector position $(x,y,z)$ in the world frame, four for the wrist orientation as a quaternion $(q_x,q_y,q_z,q_w)$, and one for the gripper aperture $g$ which sampled at $25$\,Hz (so successive timestamps are spaced $0.04$\,s apart) over the entire episode to align with the video frame rate. We therefore convert the per-frame signals into a compact text narrative aligned with the keyframe phases produced by CPD, so that the downstream reasoning stage can attend to discrete physical events instead of long numerical traces. Once an episode has been segmented into phases, each phase \([t_s,t_e]\) is captioned independently for each arm by a deterministic ruleset operating on three complementary signals: \emph{gripper aperture} (the state of the end-effector itself), \emph{end-effector translation} (where the arm went), and \emph{end-effector rotation} (how the wrist re-oriented). The three modules are decoupled and concatenated into a single per-arm paragraph, giving the VLM one self-contained line per phase per arm.

\paragraph{Gripper Aperture.}
This module consumes only the $g$ channel of the 8D state. For each arm, we read the aperture value $g$ and its temporal
derivative $v_g$ and assign every frame a discrete
\emph{stage} label,
\begin{equation}
\texttt{kind}(t)=
\begin{cases}
\textsc{open}  & v_g(t)>+\tau_v,\\
\textsc{close} & v_g(t)<-\tau_v,\\
\textsc{hold}  & \text{otherwise},
\end{cases}    
\end{equation}
where the velocity threshold $\tau_v{=}2$\,cm/s is obtained from our observation on 50 calibration episodes disjoint from FailTime-Short (stationary $|v_g|$ exceeds $2$\,cm/s in fewer than $0.01\%$ of timestamps, yet actual grasp events peak far above $2$\,cm/s). A higher threshold also keeps the caption short by avoiding over-segmentation. We group consecutive timestamps sharing the same kind into a \emph{stage}, the basic narrative unit: a stage is a maximal contiguous run of timestamps whose label is the same, indexed by its first timestamp $i$ and last timestamp $j$ (with timestamps $t_i,t_j$ and aperture values $g_i,g_j$). Two short consolidation passes then merge any neighbouring same-kind stages produced by brief sub-threshold dips and smooth out two-timestamp back-and-forth flips, so that the aperture trajectory within a phase is captioned as a compact sequence of stage-level transitions. Each resulting stage is rendered with a fixed template that always exposes both the start and end aperture, so that slow servo creep within a \textsc{hold} stage remains visible to the model:
\begin{itemize}
\itemsep0pt
\item \textsc{open}  \(\to\) \texttt{L/R gripper opens $g_i$ cm @ $t_i$ $\to$ $g_j$ cm @ $t_j$}
\item \textsc{close} \(\to\) \texttt{L/R gripper closes $g_i$ cm @ $t_i$ $\to$ $g_j$ cm @ $t_j$}
\item \textsc{hold}  \(\to\) \texttt{L/R gripper holds $g_i$ $\to$ $g_j$ cm or $g_i$ cm (if stationary)} 
\end{itemize}
Stages within a phase are concatenated and in temporal order. This rendering keeps the exact apertures at grasp/release transitions, which downstream reasoning uses to anchor onset detection.

\paragraph{End-effector Translation.}
This module consumes the $(x,y,z)$ channels of the 8-D state. For each arm we describe the end-effector trajectory within a phase with two quantities: the net displacement along each axis (end-of-phase position minus start-of-phase position) and the total path length traced out by the trajectory. The axes are anchored to the overhead \texttt{cam\_high} view: forward / backward, left / right, up / down, so the natural-language phrasing agrees with the camera the model is also looking at. Per-axis components smaller than $1$\,cm are dropped as sensor noise; the remaining axes are listed in decreasing magnitude using their directional words, and a phase with no surviving axis is reported as essentially stationary. For example:
\par\begingroup\scriptsize\begin{verbatim}
    L arm moves 47 cm down, 14 cm left (path 56 cm)
    L arm essentially stationary (path 0 cm)
\end{verbatim}
\endgroup\par
We always report the path length so the MLLM can distinguish a truly idle phase from one in which the arm moved appreciably but ended up near its starting point.

\paragraph{End-effector Rotation.}
This module consumes the four quaternion channels
$(q_x,q_y,q_z,q_w)$ of the 8-D state. For each phase we then summarise the wrist motion with three complementary scalars: the \emph{net} rotation angle from the first to the last timestamp, the \emph{total} accumulated rotation summed across all timestamps in this phase, and a \emph{stability} score in $[0,1]$ that measures axis consistency, approaching $1$ when every step shares the same axis (a single-axis monotonic rotation) and falling toward $0$ when the axis switches direction or different steps use different axes. The contrast between net and total reveals reversals: a pour-and-return motion inflates total far above net, while a pure tilt keeps them equal. Read together with the translation summary, these three scalars let the MLLM disentangle wrist-driven actions from passively dragged ones, a small translation path combined with a large stable rotation indicates an active wrist motion such as a pour or twist, whereas a large translation path alongside rotation usually means the wrist is just being carried along by the arm sweep. A phase whose net and total are both below $5^\circ$ is reported as \texttt{holds orientation}; otherwise
the wrist is captioned with the three scalars. For example:
\par\begingroup\scriptsize\begin{verbatim}
    L arm holds orientation
    L arm wrist: net 41 deg, total 48 deg, stability 0.85
\end{verbatim}
\endgroup\par

\paragraph{Failure Example.}
Consider the sample in Fig.~\ref{fig:failure-example} (\textit{pick\_mouse\_place\_laptop}). The task instructs the left arm to grasp a mouse and place it on the laptop, and the failure is that the gripper closes on empty air. The aperture trace alone makes this self-evident: \texttt{closes\,$\ldots$-0.2\,cm} in phase~2, whereas a mouse in the gripper would keep the aperture at $\sim$5\,cm. The pipeline emits \texttt{result: failure, type: translation\_error, onset\_t: 4.10\,s}, matching the ground-truth onset of $4.05$\,s.

\input{icra/appendix/inputs/pipeline_figure}

\input{icra/appendix/inputs/failure_example_analysis}

\subsection{Proprioception-Guided VLM Inference}

As described in Sec.~\ref{sec:vlm_inference}, VLM inference proceeds in two stages, with full prompts in Fig.~\ref{fig:caption_prompt}. In Stage~1, for each keyframe the VLM captions the gripper state in the wrist views and the object layout in the overhead view, reporting only what vision uniquely resolves (e.g., whether a closed gripper holds an object) and ignoring quantities which can be given by the proprioceptive narrative. The reference vocabulary (\texttt{<task\_objects>}(e.g. bowl, knife, cube,...) , \texttt{<predicates>})(e.g. cut, pour, grasp,...)  are generated by an LLM from the task description (\texttt{task\_desc}) and , once per task, and reused across all episodes as a soft grounding reference. In Stage~2, treating the sensor timestamps and apertures as authoritative, the VLM reconciles the proprioceptive dynamics with the Stage-1 scene relationships, flags any vision--sensor contradiction rather than averaging, and emits a structured diagnosis (result, onset time, reason, type, avoidance).

\input{icra/appendix/inputs/VLM_inference_prompt}

\subsection{Reflection Learning}
\label{sec:reflection learning}
Figure~\ref{fig:ICL_pipeline} shows the reflection-learning pipeline of ProTracer. The predicted failure result is compared against the ground truth; whenever the verdict is incorrect, the type is mismatched, or the onset error exceeds 0.5\,s, the prediction and ground truth are combined with the full-context history and passed to the next round. The model then outputs an edit action---\textsc{Add}, \textsc{Update}, \textsc{Merge}, or \textsc{Compress}, together with a lesson; a lesson is committed to the experience pool only after a retry on the same sample succeeds. The loop runs for 3 rounds over all 50 training samples from ViFailback training set, accumulating 11 experiences in total after 3 rounds. Fig.~\ref{fig:experiences} shows the accumulated experiences after the 3 rounds reflection learning.

\input{icra/appendix/inputs/reflection_prompt}

%% file: icra/appendix/inputs/cpd_demo.tex
\begin{figure}[h]
    \centering
    \includegraphics[width=\linewidth]{figures/demo cpd.png}
    \caption{\textbf{Visualization of bimanual proprioceptive signals for two representative episodes.} For each arm (left/right), we plot three temporal signals: gripper aperture, gripper velocity (green for opening, red for closing), and end-effector motion magnitude. Vertical lines mark detected phase boundaries, each corresponding to a key frame (kf), and shaded backgrounds delineate original un-merged phases.}
    \label{fig:cpd_demo}
\end{figure}

%% file: icra/appendix/inputs/primitive_alignment.tex
\begin{figure}[h]
    \centering
    \includegraphics[width=\linewidth]{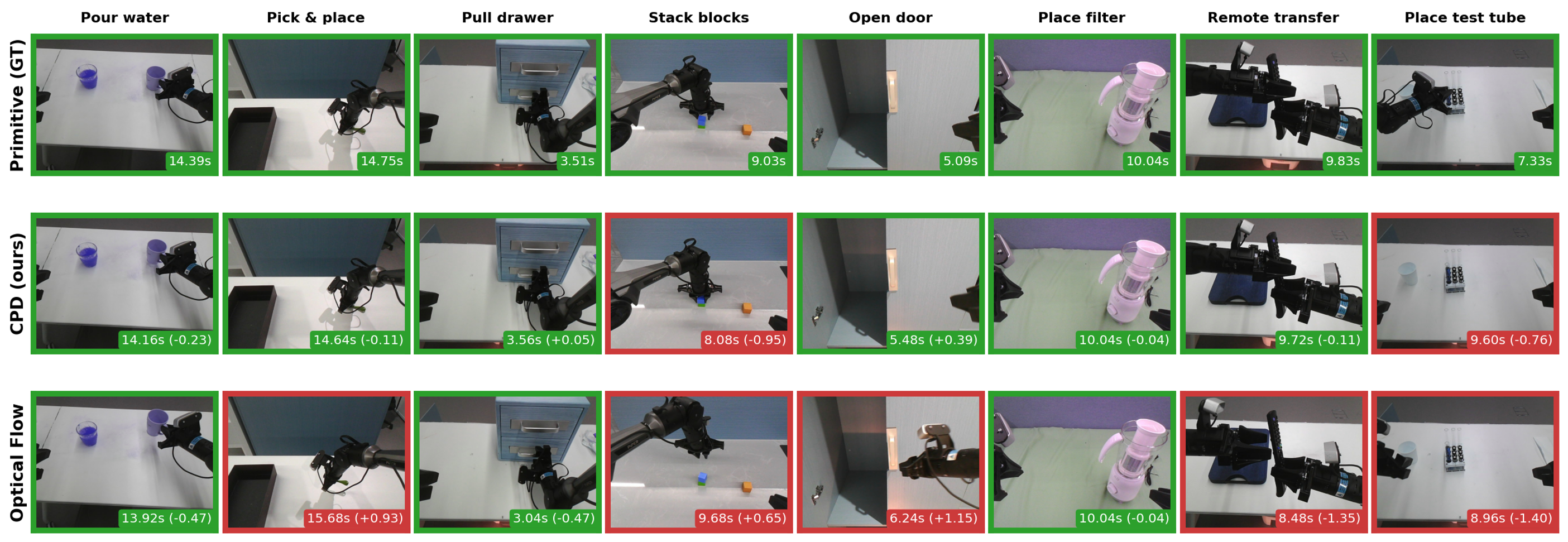}
        \caption{Qualitative comparison of keyframe selection between our CPD-based method and an optical-flow baseline across different manipulation tasks. The top row (\emph{Primitive (GT)}) shows the ground-truth primitive keyframes; the middle and bottom rows show the frames selected by CPD (ours) and optical flow, respectively. Each cell reports the selected timestamp, with the deviation from the ground-truth time (in seconds) shown in parentheses.}
        \label{fig:primitive_alignment}
\end{figure}

%% file: icra/appendix/inputs/pipeline_figure.tex
\begin{figure}[!t]
    \centering
    \includegraphics[width=\columnwidth]{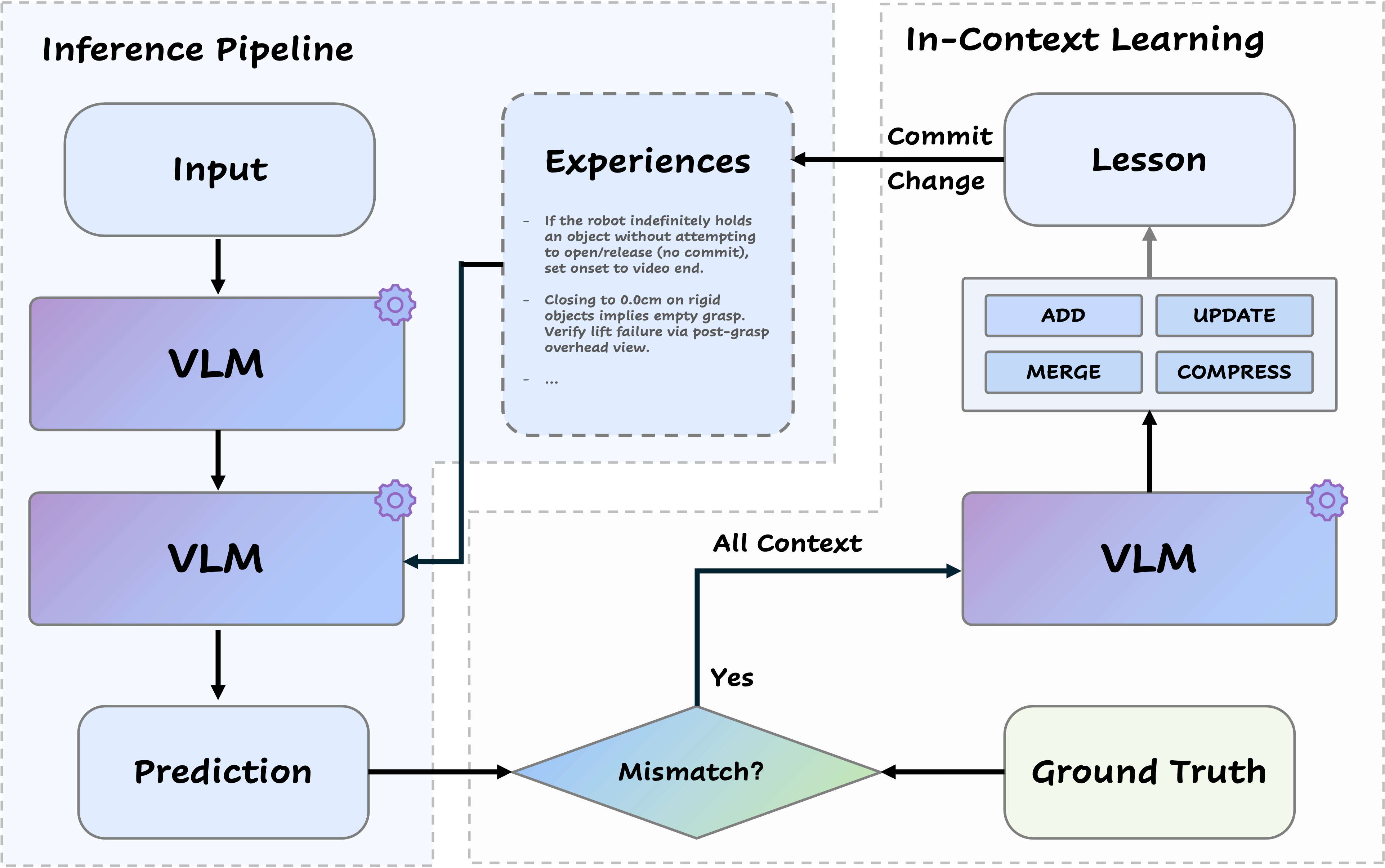}
    \caption{\textbf{The inference and learning pipeline for ProTracer.}} 
    \label{fig:ICL_pipeline}
\end{figure}

%% file: icra/appendix/inputs/failure_example_analysis.tex
\begin{figure*}[t]
\centering
\begin{tcolorbox}[colback=gray!8, colframe=black, boxrule=0.6pt, arc=0pt,
    left=8pt, right=8pt, top=6pt, bottom=4pt, width=\textwidth]
\scriptsize

\textbf{Failure example} (\textit{pick\_mouse\_place\_laptop\_episode\_10}).
\textbf{Task:} \emph{``Use the left arm to grasp the mouse and then place it on the middle of the laptop.''}
\textbf{Ground truth:} \texttt{type=translation\_error}, \texttt{onset=4.05\,s}.

\vspace{4pt}\hrule\vspace{4pt}

\ttfamily\scriptsize
\textbf{\# Phase 1: [0.00, 1.52]\,s}\\
L gripper opens 0.0\,cm @ 0.56s $\rightarrow$ 9.9\,cm @ 1.52s\\
L arm moves 17\,cm forward, 7\,cm down, 2\,cm right (path 21\,cm)\\
L arm wrist: net 16\,deg, total 26\,deg, stability 0.61\\[3pt]

\textbf{\# Phase 2: [1.52, 5.08]\,s}\\
L gripper closes 7.7\,cm @ 4.08s $\rightarrow$ \textcolor{red!75!black}{\bfseries-0.2\,cm} @ 4.92s\\
\hspace*{2em}\textcolor{red!75!black}{\itshape closes all the way: no object held}\\
L arm moves 22\,cm forward, 17\,cm down, 5\,cm right (path 36\,cm)\\
L arm wrist: net 6\,deg, total 50\,deg, stability 0.14\\[3pt]

\textbf{\# Phase 3: [5.08, 6.16]\,s}\\
L gripper holds at -0.3\,cm\\
L arm moves 14\,cm up, 12\,cm backward, 11\,cm right (path 27\,cm)\\

...
\end{tcolorbox}
\caption{A failure example analysis from signal narrative.}
\label{fig:failure-example}
\vspace{-10pt}
\end{figure*}

%% file: icra/appendix/inputs/VLM_inference_prompt.tex
\begin{figure*}[!t]
\centering
\begin{tcolorbox}[colback=gray!8, colframe=black, boxrule=0.6pt, arc=0pt,
    left=8pt, right=8pt, top=6pt, bottom=6pt, width=\textwidth]
\scriptsize

\textbf{VLM stage-1 prompt: visual captioning.}

\smallskip
I will show you $\langle$\texttt{n\_boundaries}$\rangle$ keyframe timestamps from one bimanual manipulation episode ($\langle$\texttt{n\_imgs}$\rangle$ images total). \\

\textbf{Task Description: }$\langle$\texttt{task\_desc}$\rangle$. \\
\textbf{Key Frames: }$\langle$\texttt{imgs}$\rangle$. \\

Each timestamp has up to three views, in order:
\begin{itemize}[leftmargin=1.3em, itemsep=1pt, topsep=2pt, parsep=0pt]
  \item \texttt{L = cam\_left\_wrist} --- close-up on the \emph{left} gripper's fingers.
  \item \texttt{H = cam\_high} --- overhead view of the full workspace.
  \item \texttt{R = cam\_right\_wrist} --- close-up on the \emph{right} gripper's fingers.
\end{itemize}
\medskip

\textbf{What to describe.}\;
Proprioceptive data already covers gripper aperture, arm pose, and per-phase kinematics; do \emph{not} report these. Focus on what vision uniquely resolves:
\begin{itemize}[leftmargin=1.3em, itemsep=1pt, topsep=2pt, parsep=0pt]
  \item \textbf{Wrist views:} gripper \emph{state} --- open, closed on empty space, or actually holding an object?
  \item \textbf{Overhead view:} object \emph{layout} and inter-object relations.
  \item \textbf{Configuration cues:} task-specific visual evidence.
\end{itemize}
\medskip

\textbf{Naming and vocabulary} (reference, not strict).\;
For task-relevant objects, prefer the canonical object names from \texttt{<task\_objects>}; use \texttt{left\_gripper}\,/\,\texttt{right\_gripper} for the two arms. For spatial and contact relations, \texttt{<predicates>} is a suggested vocabulary --- use it when it fits, otherwise describe in your own words. Background props may be
named freely.
\medskip

\textbf{Output format.}\;
For each timestamp, emit a block matching this template. Items in [square brackets] are placeholders you fill in:
\begin{verbatim}
<caption t="T.TTs">
- L: [left gripper at this instant]
- H: [object layout and dominant scene state]
- R: [right gripper at this instant]
- relations:
    - [subject] [predicate] [object]
</caption>
\end{verbatim}
\textbf{Constraints.}\;
Omit any \texttt{L\,/\,H\,/\,R} bullet whose view is absent at that timestamp, and omit the \texttt{relations} block when nothing is in salient contact or alignment. Describe a single instant per cell: do \emph{not} speculate about motion, intent, success/failure, or what happens between frames.

\end{tcolorbox}

\begin{tcolorbox}[colback=gray!8, colframe=black, boxrule=0.6pt, arc=0pt,
    left=8pt, right=8pt, top=6pt, bottom=6pt, width=\textwidth]
\scriptsize 

\textbf{VLM Stage 2 prompt: failure diagnosis.}
\\

You have just produced per-$(t,\text{view})$ visual captions and relations. Now produce the final judgment by combining the proprioceptive signal with the failure taxonomy. The signal describes \emph{timing and kinematics} (when, how much); vision describes the \emph{final state} (whether the target reached its goal). Treat the signal narrative as authoritative for dynamics: when a visual caption disagrees, side with the signal or explicitly flag the contradiction.
\\

\textbf{Inputs:} expected \texttt{<task\_substages>}, the per-phase  \texttt{<signal\_narrative>}, which generated deterministically from the proprioceptive signal, along with a short \texttt{<narrative\_explanation>} for the 3 dimension description, and the \texttt{<failure\_taxonomy>}.
\\

\textbf{Outputs (two steps, in order):}
\begin{itemize}[leftmargin=1.3em, itemsep=3pt, topsep=4pt, parsep=0pt]
  \item \textbf{Conclusion:} one paragraph reconciling the per-$(t,\text{view})$
  captions with the signal narrative. Walk through the keyframe phases in order and reference signal timestamps explicitly. Flag any vision--signal contradiction, and name any substage that is skipped, performed out of order, or never completed. Before declaring \textsc{success}, quote the final overhead view caption and verify the target is visibly in its goal state, proprioceptive signals alone do not prove the world visual state.
  \item \textbf{Diagnosis:} output one JSON object matching the schema below. If \texttt{result} is \texttt{success}, omit \texttt{type}, \texttt{onset\_t}, \texttt{reason}, and \texttt{avoidance}; otherwise set \texttt{onset\_t} using the per-type onset rule defined in the taxonomy. Signal narrative timestamps take priority over visual time estimates. Apply the per-type onset rule from the taxonomy above.
\end{itemize}

\begin{verbatim}
{
  "result":   "success"| "failure",
  "onset_t":  <float, seconds>,
  "reason": {
    "expected": "<intended action in the failing substage>",
    "observed": "<what happened at onset, in physical terms>",
    "result":   "<visible downstream consequence>"
  },
  "type": "[wrong_sequence| wrong_object| wrong_gripper| wrong_placement| closing_error| open_error| 
           translation_error| orientation_error| avoidable_intervention| unavoidable_intervention]",
  "avoidance": "[smallest physical action before onset_t that closes the gap]"
}
\end{verbatim}

\textbf{Experiences: }$\langle$\texttt{accumulated\  experiences}$\rangle$. 
\end{tcolorbox}
\caption{\textbf{VLM inference prompt.}}
\vspace{-20mm}
\label{fig:caption_prompt}
\end{figure*}

%% file: icra/appendix/inputs/reflection_prompt.tex
\begin{figure*}[t]
\centering
\begin{tcolorbox}[colback=gray!8, colframe=black, boxrule=0.6pt, arc=0pt,
    left=8pt, right=8pt, top=6pt, bottom=6pt, width=\textwidth]
\scriptsize
\textbf{VLM reflection learning prompt}

\smallskip
You can see your Stage-2 verdict and full reasoning above; below is the
ground-truth correct answer. Reflect on why your earlier decision diverged: which observation, signal, or reasoning step did you miss or get wrong? Then suggest a \emph{transferable} rule your future self should apply on similar episodes so the mistake does not repeat.
\\

\textbf{Priority.}\;
Focus on the biggest gap first: \textbf{verdict} (success/failure) $>$
\textbf{onset} ($> 0.5 s$) $>$ \textbf{type}. If the verdict itself is wrong, write a rule about \emph{failure detection}, not type labeling --- a type rule cannot fix a detection miss, since the type-classification step is never reached.
\\

\textbf{Inputs:} the ground-truth (\texttt{<ground\_truth\_json>}, from a human annotator, authoritative) and the rules already in your buffer (\texttt{<accumulated\_experiences>}).
\\

\textbf{Task.}\;
Write a brief introspection (1--2 sentences on what specifically you missed),
then a suggestion in \emph{one} of these formats:
\begin{verbatim}
- ADD: <new transferable rule, <=40 words>
- UPDATE <i>: <better wording of rule #i, <=40 words>
- MERGE [<i>, <j>]: <merged rule, <=40 words>
- COMPRESS
\end{verbatim}

If accumulated experience buffer exceeds the 300-word budget, re-emit the experience as a bulleted list, \texttt{COMPRESS} to make sure it $\le$300 words total, each bullet $\le$20 words. Order: most-applicable / most-general rules first; narrower or task-flavored rules last (though no rule may name a task).

Output the compressed list:

\begin{verbatim}
- <rule 1>
- <rule 2>
- ...
\end{verbatim}

Use \texttt{NONE} if no transferable rule applies or the buffer already covers it. Avoid task names, episode IDs, or specific seconds, and rules that merely restate a taxonomy name in different words.

\end{tcolorbox}
\caption{\textbf{VLM reflection learning prompt.} After each learning episode, the VLM compares its Stage-2 diagnosis against the ground truth and proposes a transferable rule (ADD / UPDATE / MERGE / COMPRESS) that is accumulated into the experience buffer.}
\label{fig:lesson_prompt}
\end{figure*}

\begin{figure*}[t]
\centering
\begin{tcolorbox}[colback=gray!8, colframe=black, boxrule=0.6pt, arc=0pt,
    left=8pt, right=8pt, top=6pt, bottom=6pt, width=\textwidth]
\scriptsize
\textbf{Accumulated experiences (reflection learning).}
\begin{itemize}[leftmargin=1.3em, itemsep=2pt, topsep=4pt, parsep=0pt]
  \item Partial aperture closing during approach (e.g.\ pre-shaping to
  ${\sim}6$\,cm) is not a commit action; set the onset at the definitive full
  close (e.g.\ near $0$\,cm) intended to secure the target.
  \item If an unrecoverable intervention occurs without a subsequent commit
  action (e.g.\ hovering), set the onset to the video end.
  \item If the robot indefinitely holds an object without attempting to open or
  release it (no commit action), set the onset to the video end.
  \item For \texttt{wrong\_gripper}, if the wrong arm contacts the target
  (e.g.\ grasping), set the onset at contact-start (just before closure begins);
  use the end of the gross approach phase only if no task-relevant contact ever
  occurs.
  \item Descending and closing an empty gripper is a \texttt{translation\_error}
  at close-start, not premature transport or a skipped grasp.
  \item Closing to $0.0$\,cm on a rigid object implies an empty grasp; verify the
  lift failure via the post-grasp overhead view.
  \item If a failed grasp stems from a visible positional offset or angular
  misalignment relative to the object's affordance, classify it as
  \texttt{translation\_error} or \texttt{orientation\_error} at contact-/close-start,
  rather than \texttt{closing\_error} at the subsequent lift.
  \item Inadvertently opening the gripper and losing hold of an object
  mid-transport, -alignment, or -articulation (e.g.\ pulling a door) is a
  \texttt{closing\_error}, not a \texttt{wrong\_placement}.
  \item If a human shifts a receptacle and the robot releases at the original
  location, classify it as \texttt{avoidable\_intervention} at release-start.
  \item Trust the sensor narrative for arm identification: if it shows the
  unexpected arm moving while the expected arm stays stationary, classify it as
  \texttt{wrong\_gripper} rather than assuming a sensor--vision contradiction.
  \item If the gripper stops closing at an aperture noticeably wider than the
  target's thickness (e.g.\ ${>}2$\,cm for a thin plate rim), classify it as
  \texttt{closing\_error} at the subsequent lift/transport start, not
  \texttt{translation\_error}.
\end{itemize}
\end{tcolorbox}
\caption{The experiences accumulated by ProTracer's reflection-learning
loop over the training set.}
\label{fig:experiences}
\end{figure*}

%% file: icra/appendix/evaluation.tex
\section{Details of Evaluation}

Fig.~\ref{fig:judge-reason} shows the LLM-as-judge prompt used to score the
predicted failure reason and avoidance against the ground truth. We follow the
same evaluation schema as~\cite{vifailback} as closely as possible to ensure
consistency.

\label{sec:evluation}

\begin{figure*}[t]
\centering
\begin{tcolorbox}[colback=gray!8, colframe=black, boxrule=0.6pt, arc=0pt,
    left=8pt, right=8pt, top=6pt, bottom=6pt, width=\textwidth]
\scriptsize\raggedright
\textbf{LLM-as-judge prompt — Failure Reason.}
You are an expert evaluator specializing in robotic manipulation tasks,
capable of understanding and judging the semantic accuracy of a failure
reason involving robot actions, objects, and spatial reasoning. Rate
how semantically consistent Text~B (candidate) is compared to Text~A
(reference) on a continuous scale from $0.0$ to $1.0$. Focus on whether
the two descriptions convey the same manipulation \emph{intent,
sequence, and outcome}.

\textbf{Key aspects (if applicable):}
\begin{itemize}[leftmargin=1.3em, itemsep=2pt, topsep=4pt, parsep=0pt]
  \item \textbf{Gripper usage}: left vs.\ right gripper, open/close actions.
  \item \textbf{Action correctness}: pick, place, move, align, push, lift, etc.
  \item \textbf{Object and target consistency}: same object names and corresponding targets.
  \item \textbf{Order and causality}: whether the sequence of steps is preserved.
  \item \textbf{Success condition}: whether the goal or outcome matches.
\end{itemize}

\textbf{Scoring guidance:}
\begin{itemize}[leftmargin=1.3em, itemsep=2pt, topsep=4pt, parsep=0pt]
  \item $1.0$ — Exactly the same meaning and correct execution description.
  \item $0.8$--$0.9$ — Minor paraphrasing differences but semantically identical in robot actions and outcomes.
  \item $0.6$--$0.7$ — Mostly correct but with small errors (e.g., gripper swapped, one step missing).
  \item $0.3$--$0.5$ — Partially correct but with notable mismatches in object, action, or order.
  \item $0.1$--$0.2$ — Only slightly related, mostly incorrect.
  \item $0.0$ — Completely unrelated or contradicting actions.
\end{itemize}

\vspace{2pt}
\textbf{Input:}\\
\hspace*{1.2em}\texttt{Text A (reference): \{ground\_truth\}}\\
\hspace*{1.2em}\texttt{Text B (candidate): \{model\_raw\_response\}}\\
\textbf{Output:} a single numeric score in $[0.0, 1.0]$ — no explanation.
\end{tcolorbox}

\begin{tcolorbox}[colback=gray!8, colframe=black, boxrule=0.6pt, arc=0pt,
    left=8pt, right=8pt, top=6pt, bottom=6pt, width=\textwidth]
\scriptsize\raggedright
\textbf{LLM-as-judge prompt — Avoidance.}
You are an expert evaluator for robotic manipulation tasks. Rate how
semantically consistent Text~B (candidate corrective action) is with
Text~A (reference) on a continuous scale from $0.0$ to $1.0$. Text~B
should describe a single corrective action taken \emph{before} failure
onset. Score only on dimensions explicitly present in the reference;
extra detail in the candidate is not penalised unless it directly
contradicts the reference.

\textbf{Key aspects:}
\begin{itemize}[leftmargin=1.3em, itemsep=2pt, topsep=4pt, parsep=0pt]
  \item \textbf{Gripper}: left vs.\ right (required if in reference).
  \item \textbf{Action verb}: close, open, move, rotate, align, etc.\ — must match in kind.
  \item \textbf{Object} (only if named in reference): same target.
  \item \textbf{Direction or magnitude} (only if present in the reference): spatial direction such as left, right, up, down, forward, or backward.
  \item \textbf{Unavoidable case}: if the reference says the failure is unavoidable, the candidate must also mention this or suggest a practical workaround (e.g.\ keeping still or waiting for human resolution).
\end{itemize}

\textbf{Scoring:}
\begin{itemize}[leftmargin=1.3em, itemsep=2pt, topsep=4pt, parsep=0pt]
  \item $1.0$ — Same meaning.
  \item $0.8$--$0.9$ — Minor paraphrasing, semantically identical.
  \item $0.6$--$0.7$ — Same gripper + verb, but a direction/magnitude present in the reference is slightly off.
  \item $0.3$--$0.5$ — Wrong arm, related-but-distinct action, or notable mismatch.
  \item $0.1$--$0.2$ — Opposite direction or action-kind mismatch (e.g., rotate vs.\ shift).
  \item $0.0$ — Unrelated, contradicting, or unavoidable-case mismatch.
\end{itemize}

\vspace{2pt}
\textbf{Input:}\\
\hspace*{1.2em}\texttt{Text A (reference): \{ground\_truth\}}\\
\hspace*{1.2em}\texttt{Text B (candidate): \{model\_raw\_response\}}\\
\textbf{Output:} a single numeric score in $[0.0, 1.0]$ — no explanation.
\end{tcolorbox}

\caption{LLM-judge prompt used to score the predicted failure reason and avoidance against the ground-truth reason and avoidance.}
\label{fig:judge-reason}
\end{figure*}